\documentclass[10pt]{article}
\usepackage[preprint]{neurips_2026}
\usepackage[T1]{fontenc}
\usepackage[utf8]{inputenc}
\usepackage{amsfonts}
\usepackage{microtype}
\usepackage{amsmath}
\usepackage{booktabs}
\usepackage{graphicx}
\usepackage{xcolor}
\usepackage{enumitem}
\usepackage{float}
\usepackage{placeins}
\usepackage{hyperref}
\usepackage{url}
\hypersetup{colorlinks=true,linkcolor=blue!50!black,citecolor=blue!50!black,urlcolor=blue!50!black,pdftitle={Self-supervised behavior encoders rarely beat a training-free baseline},pdfauthor={Ziwen Chen}}
\setlist{nosep,leftmargin=1.5em}
\newcommand{\gap}[3]{$#1$\,[$#2$, $#3$]}
\newsavebox{\fitboxsave}
\newcommand{\fitbox}[2]{\begingroup\sbox{\fitboxsave}{#2}%
  \ifdim\wd\fitboxsave>#1\relax\resizebox{#1}{!}{\usebox{\fitboxsave}}\else\usebox{\fitboxsave}\fi\endgroup}
\newcommand{\codestatement}{The code for both variants, the frozen fits and the commands that score them with each
benchmark's own evaluator are at \url{https://github.com/uranitonium/motion-stats-baseline}.}
\newcommand{\aistatement}{The author used Claude (Anthropic), an AI assistant, to implement and run experiment code, to
search for related work and to draft and edit the text. All results were checked against the saved result files, the
encoders were checked against their published scores, and the author takes full responsibility for the content.}

\title{Self-supervised behavior encoders rarely beat a training-free baseline}
\author{Ziwen Chen\\
  University of Illinois Urbana-Champaign\\
  \texttt{ziwenc3@illinois.edu}}

\begin{document}
\maketitle

\begin{abstract}
Self-supervised encoders of animal and human pose are compared with one another, and rarely with simple summaries of the
poses they encode. We propose such a summary that needs no training: each body's keypoints and velocities in its own
frame, a small block on the nearest other animals, and statistics over windows from half a second to the whole clip,
reduced by PCA. Under the standard protocol of frozen, size-matched representations read by a linear classifier, we
compare it with seven encoders on five benchmarks spanning mice, flies and humans. It scores above hBehaveMAE and BAMS
wherever they were run, with every paired gap surviving a Holm correction, and leads from about a tenth of the labels;
only HiDRA, built for pairs of mice, beats it, on mouse pairs. Plain kinematic measurements in place of the keypoints
widen the lead over hBehaveMAE. Two choices carry the baseline: the body's own frame and windows matched to the labels' time scale.
On the real datasets, linear readouts recover about half as much motion from hBehaveMAE's embeddings as from the
baseline.
\end{abstract}

\section{Introduction}\label{sec8:intro}

Behavior analysis increasingly relies on representations learned from tracked poses without labels. Encoders such as
BAMS~\citep{azabou2023bams}, hBehaveMAE~\citep{stoffl2024hbehavemae} and MICE~\citep{su2026mice} are pretrained on a
dataset's unlabeled recordings and judged by how well a small classifier reads behavior from their frozen embeddings,
and each new encoder is compared with the previous ones. How much an untrained summary of the same poses already
provides is rarely measured, although several encoders lean on hand-built inputs (BAMS predicts histograms of future
movement features, and its mouse recipe takes hand-built pose features as input), and MARS's hand-built features, read
by a trained classifier, have been found ahead of BAMS and hBehaveMAE on CalMS21~\citep{li2026visualcontext}.

\begin{figure*}[t]
\centering
\includegraphics[width=\textwidth]{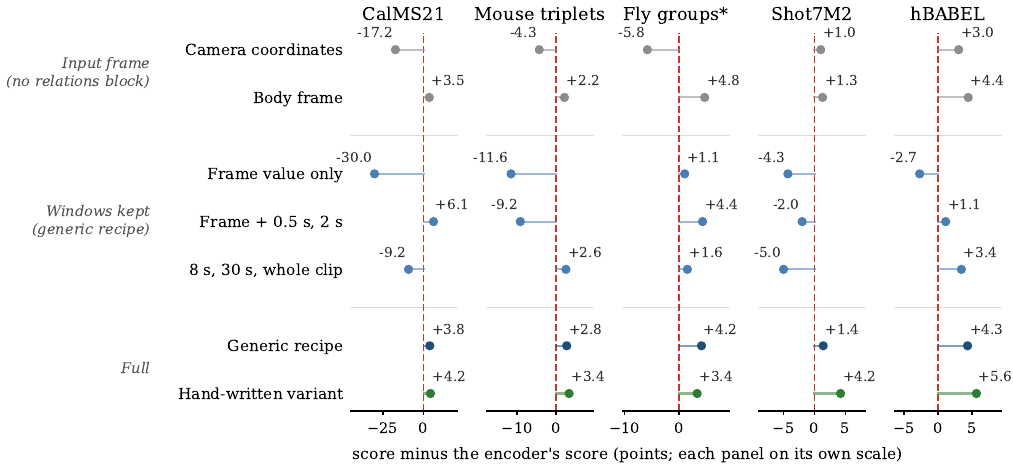}
\caption{Where the result comes from (linear classifier, all labels). Each point is a score minus hBehaveMAE (*on the flies, BAMS's published score); right of the dashed line is better than the encoder. Top: the input frame, without the relations block. Middle: the generic recipe keeping only some of its windows. Bottom: both full variants.}\label{fig8:ladder}
\end{figure*}

We measure it with a baseline that has no trained parameters and needs only minimal per-dataset choices: each body's
keypoints and velocities in its own frame, a small block on the nearest other animals, statistics over windows from
half a second to the whole clip, and PCA to the benchmark's budget. We compare it with seven self-supervised encoders on
five benchmarks spanning mice, flies and humans, under each benchmark's own protocol: frozen representations, at most the
same number of dimensions and the evaluators' linear classifier (Figure~\ref{fig8:ladder}). We find:
\begin{itemize}
\item \textbf{Under the standard protocol, encoders rarely beat the baseline.} The baseline scores above hBehaveMAE on
all four benchmarks where hBehaveMAE was run, and above BAMS on the three where we retrained it. On the flies it scores
above BAMS's published result. The lead over hBehaveMAE holds from about a tenth of the labels, across encoder seeds and
for a new CalMS21 annotator. With all labels, only HiDRA beats it, on the mouse pairs it was built for
(Section~\ref{sec8:results}).
\item \textbf{A simple add-on widens the gap.} Replacing the keypoints with a small set of plain kinematic measurements
(23 to 100 per frame, written once per benchmark and not tuned on scores) raises the lead over hBehaveMAE to 3.4 to 5.6
points. The gain over the baseline itself is resolved on the two human benchmarks.
\item \textbf{The readout shapes the reading.} A temporal head narrows several gaps, and with it even raw body-frame
keypoints, without windows, beat hBehaveMAE. Under that head, BAMS with random weights is not resolved from trained
BAMS on CalMS21, while pretraining lifts hBehaveMAE well above its random weights. We recommend reporting a training-free baseline,
a random-weight control and more than one readout with every new encoder.
\item \textbf{Two choices carry the baseline}: the body's own frame on the animal benchmarks, and windows matched to
the labels' time scale. As diagnostics, linear readouts recover about as much position from hBehaveMAE as from the
recipe but about half the motion, and on CalMS21 recoverable motion tracks an encoder's score
(Sections~\ref{sec8:needs} and~\ref{sec8:why}).
\end{itemize}

\section{Related work}\label{sec8:related}

\paragraph{Behavior encoders.} TREBA learns trajectory embeddings guided by expert-written programs~\citep{sun2021treba};
BAMS trains short- and long-range temporal convolutions to predict histograms of future actions and ranked first overall
in the MABe 2022 behavior challenge~\citep{azabou2023bams}; hBehaveMAE masks and reconstructs pose at several levels of a
hierarchy~\citep{stoffl2024hbehavemae}; LISBET pretrains a transformer on social pretext tasks~\citep{chindemi2023lisbet};
HiDRA forecasts keypoint trajectories with message passing within and between mice~\citep{marks2026doom}; and MICE is a
hierarchical foundation model for mouse behavior~\citep{su2026mice}. General-purpose encoders such as
TS2Vec~\citep{yue2022ts2vec} and CEBRA~\citep{schneider2023cebra} are applied to behavior as well, and B-SOiD, VAME and
Keypoint-MoSeq segment it into discrete motifs~\citep{hsu2021bsoid,luxem2022vame,weinreb2024kpms}.

\paragraph{Hand-built features and simple baselines.} JAABA, MARS and SimBA classify behavior from hand-built features
of pose and of the relations between animals~\citep{kabra2013jaaba,segalin2021mars,goodwin2024simba}.
\citet{li2026visualcontext} found MARS's features ahead of BAMS and hBehaveMAE on CalMS21 with a trained recurrent
classifier (87.8 macro F1 against 70.8 and 62.4); our baseline has no behavior-specific engineering. Which side wins
depends on the readout: fine-tuned contrastive pose features beat hand-crafted ones on
CalMS21~\citep{zhou2022contrastivepose}, and the video encoder BEAST surpasses SimBA's features with a temporal
classifier, and its linear readouts trail its nonlinear ones~\citep{wang2025beast}. Elsewhere, simple baselines have reset
expectations: linear models rival Transformer forecasters~\citep{zeng2023dlinear}, random convolutional kernels matched
state-of-the-art time-series classification~\citep{dempster2020rocket}, and careful re-evaluation shrank reported gains
in metric learning, language modeling and recommendation~\citep{musgrave2020reality,melis2018state,dacrema2019progress}.
We also compare with MiniRocket and catch22~\citep{dempster2021minirocket,lubba2019catch22}.

\section{A training-free baseline}\label{sec8:method}

The baseline, which we call the generic recipe (the recipe for short), turns tracked keypoints into a fixed number of values per frame. It has
no trained parameters: the normalizing moments, the histogram edges and a PCA are fitted without labels and then frozen.
By training-free we mean exactly this: no neural network is trained; the only label-dependent step is the choice among
four layouts by cross-validation on training labels, as each encoder's version is chosen.

\paragraph{Each body in its own frame.} For every body we take its keypoints and their velocities (per second) in a
frame centered on the body and turned to its heading: tail or abdomen to head for animals, the hip line for human
skeletons. Lengths are in body lengths (CalMS21, the human skeletons) or the dataset's fixed unit (millimetres for the
flies, 100 pixels for the triplets' fixed camera); tracking gaps are interpolated, and the CalMS21 and fly keypoints are
median-filtered over a few frames.

\paragraph{The nearest other animals.} For groups, one block, the same for every species, adds the distances from each
animal's center to its two nearest others, the nearest one's position and velocity in the animal's own frame, and the
closing speed (on CalMS21 both distances refer to the one other mouse). Per-animal numbers are sorted over the three
triplet mice and summarized by mean, spread, minimum and maximum over up to 11 flies, so the result does not depend on
animal order; the two CalMS21 mice keep the roles the benchmark gives them.

\paragraph{Windows and reduction.} Each number contributes its value at the frame and, over centered windows of 0.5,
2, 8 and 30\,s and, where a benchmark has clips, episodes or sequences, the whole clip, its mean, standard deviation and
an 8-bin histogram with edges at training quantiles. The statistics are standardized and reduced by PCA to the
benchmark's budget (Table~\ref{tab8:bench}). One of four layouts (a single PCA or one each for short and long windows,
each with or without the histograms) is chosen by cross-validation on training labels, as each encoder's variant is.
Fitting takes minutes on a CPU (Appendix~\ref{app:recipe}).

\paragraph{An optional add-on: hand-written measurements.} Where a few plain kinematic quantities are easy to write
down, they can replace the keypoints as the per-frame input, with the same windows and reduction; we call this the
hand-written variant. On CalMS21 its 23 measurements are the distance between the mice and whether they touch, each
mouse's speed and its speed toward the other, the closing speed, how still each mouse is, each nose's distance to the
other mouse, whether each head points at the other and how fast it turns, and each body's length; the human skeletons
get joint angles, limb positions and speeds, and heights. Each set (23 to 100 measurements) is under 100 lines of
Python, none defines a behavior, none was tuned, and each was fixed before any test score on its benchmark existed.

\section{Experimental setup}\label{sec:setup}

\begin{table}[t]
\centering\small
\caption{Benchmarks, their labels and the budget of numbers per frame every method gets.}\label{tab8:bench}
\setlength{\tabcolsep}{2.5pt}
\fitbox{\linewidth}{%
\begin{tabular}{@{}llcl@{}}
\toprule
Benchmark & Bodies & Budget & Labels (time scale) \\
\midrule
CalMS21~\citep{sun2021calms21} & 2 mice & 128 & 3 social behaviors (bouts) \\
MABe22 triplets~\citep{sun2023mabe22} & 3 mice & 128 & strain, lighting (clip) \\
MABe22 flies~\citep{sun2023mabe22} & $\le$11 flies & 256 & 50 tasks, mostly group \\
Shot7M2~\citep{stoffl2024hbehavemae} & 1 simulated body & 64 & activity, action, moveme \\
hBABEL~\citep{stoffl2024hbehavemae} & 1 human (mocap) & 64 & 180 actions (frame, sequence) \\
\bottomrule
\end{tabular}}
\end{table}

\paragraph{Benchmarks.} CalMS21 Task 1 labels attack, investigation and mount in pairs of
mice~\citep{sun2021calms21}. The MABe22 mouse triplets~\citep{sun2023mabe22} label whole clips by the mice's strain and
whether the lights are on, and the MABe22 fly groups by 50 tasks, most of them traits of the group such as genotype;
these two test what a representation keeps about the animals and their conditions rather than moment-to-moment
behavior. Shot7M2 is a simulated basketball player labeled with activities, actions and movemes, and hBABEL is real
human motion capture labeled with 180 action tasks~\citep{stoffl2024hbehavemae}. We built hBABEL's files from AMASS and
BABEL~\citep{mahmood2019amass,punnakkal2021babel} with the authors' scripts; on them the released hBehaveMAE reproduces the score
computed from its published columns (16.95 against 16.98; Appendix~\ref{app:hbabel}).

\paragraph{One protocol for every method.} Every representation is frozen, and none gives the classifier more numbers
per frame than the budget. Every method is read by the classifier that the MABe22, Shot7M2 and hBABEL evaluators train
on frozen embeddings: a ridge classifier with balanced class weights and a penalty of 1. On CalMS21 we fit it on the 70
training videos and score the 19 test videos once, by macro F1 over the three behaviors. Where a method has several
variants (hBehaveMAE's levels, our layouts, BAMS's time shifts), cross-validation on training videos picks one.

Some choices favor the encoder. The released hBehaveMAE is reported at its best single level on the test set, as its
authors report it (on CalMS21 the level is chosen by cross-validation), and its own script reduces wider levels with a
PCA fitted on frames that include the test videos. Other choices favor the recipe: it gets body-frame velocities and,
for groups of animals, the relations block, which the encoders' keypoint inputs lack (Section~\ref{sec8:needs}
quantifies both). Both sides see later frames. The recipe's centered windows reach up to 15\,s ahead, and its
whole-clip windows the rest of the clip. hBehaveMAE's script embeds every 30\,s window at stride one and averages, for
each frame, over all windows that contain it, so each frame's embedding draws on up to 30\,s before and after it.
Removing the future from the recipe alone is therefore a one-sided handicap rather than a fairness check; we report it
in Appendix~\ref{app:robust}.

\paragraph{How the recipe was fixed.} Its window statistics, histograms, standardization, PCA and four layouts are the
hand-written variant's, fixed before any test score in this study. Its inputs are body-frame keypoints with velocities,
which had been scored as ablations of the hand-written variant, and the relations block. The block was added, as one
block for all three animal benchmarks, after a keypoint-only version trailed the hand-written variant on the flies; on
the flies its own effect is not resolved (Section~\ref{sec8:needs}). On every benchmark the layout is the one
cross-validation picks on training labels. Once the recipe was defined with the block, no setting was changed after its
test scores, and the released code differs from the code that produced them only in comments. The encoders were
likewise developed on these benchmarks. hBehaveMAE was introduced together with Shot7M2 and hBABEL and is reported at
its best level on each~\citep{stoffl2024hbehavemae}; BAMS ranked first in the MABe 2022 challenge, whose mouse and fly
data we use~\citep{azabou2023bams}; and HiDRA was pretrained on videos that include CalMS21 and
MABe22~\citep{marks2026doom}.

\paragraph{Encoders.} On CalMS21 we retrained BAMS and hBehaveMAE with their authors' code on the unlabeled and
training videos. BAMS's mouse recipe takes hand-built features, so we adapted its code to take both mice's keypoints in
the resident's body frame, scaled by their median and interquartile range so that rare tracking jumps do not set its
input scale. Read with a one-hidden-layer MLP probe, as MICE reads it, it scores 72.7, close to the 75.0 that MICE
reports for BAMS on CalMS21~\citep{su2026mice}. Elsewhere we use the released hBehaveMAE checkpoints with the authors'
evaluators, and BAMS retrained by us on body-frame keypoints on Shot7M2 and hBABEL, where it scores above BAMS's
published results (57.3 against 53.7, and 14.5 against 10.3). On the flies we use BAMS's published score. We also ran
TS2Vec and CEBRA-Time everywhere, the released TREBA, LISBET and HiDRA on CalMS21, and HiDRA on the triplets. HiDRA's
released trunks were pretrained on 9,008 videos that include CalMS21 and MABe22 (Appendix~\ref{app:protocol}).

\paragraph{Uncertainty.} Intervals are 95\% intervals from 10,000 paired resamples of test videos, clips, episodes or
sequences. We call a gap whose interval includes zero not resolved, rather than a tie, since such an interval does not
show the two are equal. The 14 gaps in the bottom rows of Table~\ref{tab8:main} are the primary comparisons, corrected
together with Holm's method; every other comparison, including the label curves and the correlation of
Section~\ref{sec8:why}, is exploratory and uncorrected. Every primary gap whose interval excludes zero survives the
correction; HiDRA's lead on CalMS21 does so with the permutation test, which we use on CalMS21
(Appendix~\ref{app:multiplicity}). hBehaveMAE was trained with three seeds on CalMS21, and the recipe leads each.
Resampling the seeds together with the test videos gives \gap{+4.8}{+2.7}{+8.5} for the mean over seeds, close to the
video-only interval of [$+3.0$, $+8.4$], so the 19 test videos dominate the uncertainty. TS2Vec and CEBRA-Time have extra
seeds on CalMS21 (Appendix~\ref{app:robust}); the other encoders are single runs or released checkpoints, so their
intervals hold the weights fixed. Attack occurs in only 5 test videos, but scored on investigation and mount alone the
recipe still leads hBehaveMAE, by 2.6 points. Bias-corrected intervals and permutation tests leave every
reading unchanged (Appendix~\ref{app:robust}).

\section{Results}\label{sec8:results}

\begin{table}[t]
\centering\small
\caption{Holm-adjusted $p$ for the 14 primary gaps of Table~\ref{tab8:main}.}\label{tab:holm}
\setlength{\tabcolsep}{3pt}
\fitbox{\linewidth}{%
\begin{tabular}{@{}llcc@{}}
\toprule
Benchmark & Gap & Value & Holm $p$ \\
\midrule
CalMS21 & Generic minus hBehaveMAE & $+3.8$ & 0.007 \\
 & Generic minus BAMS & $+23.2$ & 0.007 \\
 & Generic minus HiDRA & $-3.5$ & 0.007 \\
 & Hand-written minus generic & $+0.4$ & 1 \\
Mouse triplets & Generic minus hBehaveMAE & $+2.8$ & 0.003 \\
 & Generic minus HiDRA & $+7.3$ & 0.0014 \\
 & Hand-written minus generic & $+0.6$ & 1 \\
Fly groups & Hand-written minus generic & $-0.8$ & 1 \\
Shot7M2 & Generic minus hBehaveMAE & $+1.4$ & 0.0014 \\
 & Generic minus BAMS & $+8.7$ & 0.0014 \\
 & Hand-written minus generic & $+2.8$ & 0.0014 \\
hBABEL & Generic minus hBehaveMAE & $+4.3$ & 0.0014 \\
 & Generic minus BAMS & $+6.4$ & 0.0014 \\
 & Hand-written minus generic & $+1.3$ & 0.007 \\
\bottomrule
\end{tabular}}
\end{table}

\begin{table*}[t]
\centering\small
\caption{Main comparison, all labels, linear classifier, in each benchmark's own score. Bottom rows: paired gaps with 95\% intervals. --: not run. $^\ast$Pretrained on data that include CalMS21 and MABe22. $^\dagger$Published score, no paired interval. $^\ddagger$Best single level, as published.}\label{tab8:main}
\setlength{\tabcolsep}{4pt}
\fitbox{\textwidth}{%
\begin{tabular}{@{}lccccc@{}}
\toprule
 & CalMS21 & Mouse triplets & Fly groups & Shot7M2 & hBABEL \\
\midrule
Generic recipe (ours) & 79.7 & 83.1 & \textbf{52.4} & 66.0 & 20.9 \\
Hand-written variant (ours) & 80.1 & \textbf{83.7} & 51.6 & \textbf{68.8} & \textbf{22.1} \\
\midrule
hBehaveMAE & 75.9 & 80.3 & -- & 64.7$^\ddagger$ & 16.5 \\
BAMS & 56.6 & -- & 48.2$^\dagger$ & 57.3 & 14.5 \\
TS2Vec & 43.8 & 59.3 & 39.5 & 56.6 & 15.2 \\
CEBRA-Time & 49.5 & 64.9 & 38.2 & 57.5 & 11.4 \\
MiniRocket, body-frame keypoints & -- & -- & 47.2 & 62.1 & 20.0 \\
catch22, body-frame keypoints & -- & -- & 42.7 & 60.7 & 19.8 \\
LISBET, released & 74.9 & -- & -- & -- & -- \\
HiDRA, released$^\ast$ & \textbf{83.2} & 75.8 & -- & -- & -- \\
\midrule
Generic minus hBehaveMAE & \gap{+3.8}{+2.0}{+7.4} & \gap{+2.8}{+1.5}{+4.0} & -- & \gap{+1.4}{+0.8}{+1.9} & \gap{+4.3}{+3.4}{+5.1} \\
Generic minus BAMS & \gap{+23.2}{+19.4}{+31.8} & -- & $+4.2^\dagger$ & \gap{+8.7}{+8.3}{+9.3} & \gap{+6.4}{+5.4}{+7.1} \\
Generic minus HiDRA$^\ast$ & \gap{-3.5}{-4.7}{-1.5} & \gap{+7.3}{+5.9}{+8.6} & -- & -- & -- \\
Hand-written minus generic & \gap{+0.4}{-1.8}{+1.7} & \gap{+0.6}{-0.7}{+1.9} & \gap{-0.8}{-3.1}{+1.7} & \gap{+2.8}{+2.5}{+3.1} & \gap{+1.3}{+0.5}{+2.1} \\
\bottomrule
\end{tabular}}
\end{table*}

\begin{figure*}[t]
\centering
\includegraphics[width=\textwidth]{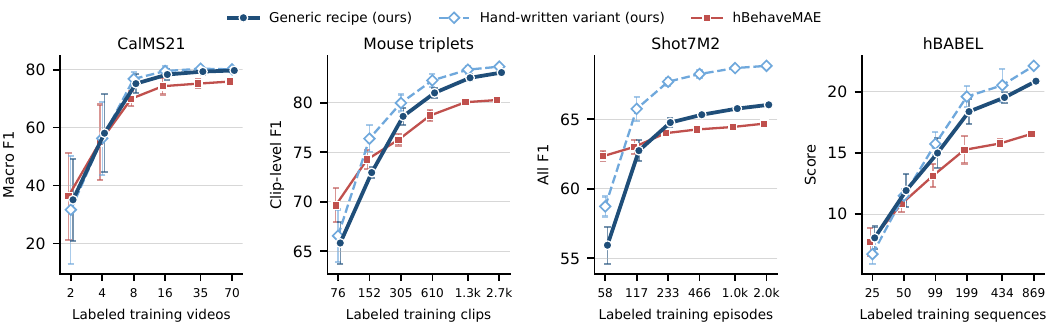}
\caption{Score against the number of labeled training videos, clips, episodes or sequences; points average ten fixed (CalMS21) or five random subsets, bars show one standard deviation.}\label{fig8:labels}
\end{figure*}

\begin{table*}[t]
\centering\small
\caption{Learned readouts, all labels: the generic recipe minus each encoder (top) and the hand-written variant minus hBehaveMAE (bottom), with paired 95\% intervals. $^\ast$Pretrained on data that include CalMS21. --: not run.}\label{tab8:robust}
\setlength{\tabcolsep}{4pt}
\fitbox{\linewidth}{%
\begin{tabular}{@{}llccc@{}}
\toprule
Benchmark & Minus & Ridge (the protocol) & MLP probe & Temporal head \\
\midrule
\multicolumn{5}{@{}l}{\emph{Generic recipe minus each encoder}} \\
CalMS21 & hBehaveMAE & \gap{+3.8}{+2.0}{+7.4} & \gap{+3.4}{+2.5}{+4.6} & \gap{+3.4}{+2.1}{+4.8} \\
 & BAMS & \gap{+23.2}{+19.4}{+31.8} & \gap{+13.1}{+10.6}{+21.4} & \gap{-0.9}{-1.9}{+0.5} \\
 & LISBET, released & \gap{+4.9}{+2.8}{+9.3} & \gap{+5.9}{+4.5}{+8.5} & \gap{-0.9}{-2.0}{+0.3} \\
 & TREBA, released & \gap{+17.7}{+12.9}{+28.6} & \gap{+17.6}{+15.0}{+24.8} & \gap{+2.7}{+0.6}{+4.5} \\
 & CEBRA-Time, body frame & \gap{+10.1}{+5.8}{+19.5} & \gap{+11.7}{+8.8}{+20.2} & \gap{+2.0}{+0.8}{+4.5} \\
 & TS2Vec, body frame & \gap{+29.8}{+25.2}{+37.8} & \gap{+13.1}{+11.1}{+21.3} & \gap{+3.6}{+1.6}{+9.3} \\
 & HiDRA, released$^\ast$ & \gap{-3.5}{-4.7}{-1.5} & \gap{-2.3}{-3.2}{-0.8} & \gap{-4.7}{-5.7}{-3.9} \\
Mouse triplets & hBehaveMAE & \gap{+2.8}{+1.5}{+4.0} & \gap{+1.1}{-0.4}{+2.6} & -- \\
Shot7M2 & hBehaveMAE & \gap{+1.4}{+0.8}{+1.9} & \gap{-3.7}{-4.0}{-3.3} & \gap{-3.2}{-3.6}{-2.8} \\
hBABEL & hBehaveMAE & \gap{+4.3}{+3.4}{+5.1} & \gap{+5.9}{+4.3}{+7.1} & \gap{+6.0}{+4.3}{+7.2} \\
\midrule
\multicolumn{5}{@{}l}{\emph{Hand-written variant minus hBehaveMAE}} \\
CalMS21 & hBehaveMAE & \gap{+4.2}{+2.4}{+6.3} & \gap{+3.4}{+2.7}{+4.5} & \gap{+3.9}{+2.3}{+5.1} \\
Mouse triplets & hBehaveMAE & \gap{+3.4}{+2.1}{+4.7} & \gap{+3.2}{+1.7}{+4.6} & -- \\
Shot7M2 & hBehaveMAE & \gap{+4.2}{+3.7}{+4.6} & \gap{-0.8}{-1.1}{-0.4} & \gap{-0.3}{-0.5}{+0.0} \\
hBABEL & hBehaveMAE & \gap{+5.6}{+4.6}{+6.4} & \gap{+5.8}{+4.2}{+7.3} & \gap{+4.2}{+2.8}{+5.2} \\
\bottomrule
\end{tabular}}
\end{table*}

\paragraph{With all labels.} Under the linear protocol the recipe matches or beats hBehaveMAE and BAMS, the encoders
pretrained on each benchmark's own data, everywhere (Table~\ref{tab8:main}). Its largest margin is over BAMS on CalMS21
(23.2 points). BAMS's embedding of a frame sees only the past, but even with past-only windows the recipe still leads
it by 19.2 (\gap{+19.2}{+16.0}{+25.3}). Against hBehaveMAE, whose embeddings also average over later frames, past-only
windows are a one-sided handicap; both comparisons are in Appendix~\ref{app:robust}. Its smallest margin is on Shot7M2,
over hBehaveMAE's best single level; there hBehaveMAE's three levels combined, its authors' option, tie the recipe, and
taking each scale from its best level puts hBehaveMAE ahead (Appendix~\ref{app:diag}). On the triplets the recipe leads
the released hBehaveMAE, although that model was pretrained on the test clips; this score covers strain and lighting,
and on the triplets' nine frame-level behaviors the two are close (Appendix~\ref{app:benchmarks}). On the flies the
recipe's interval lies above BAMS's published score. TS2Vec, CEBRA-Time, MiniRocket and catch22 trail it wherever they
were run (Appendix~\ref{app:carries}).

\paragraph{The add-on widens the gap.} With the hand-written measurements in place of the keypoints, the lead over
hBehaveMAE grows on every benchmark, to 3.4 to 5.6 points. The gain over the recipe is resolved on Shot7M2 and hBABEL
(2.8 and 1.3 points) and not on the animal benchmarks. MARS's own hand-built features, through the same pipeline, score
78.7 on CalMS21, not resolved from either variant (Appendix~\ref{app:further}).

\paragraph{Encoders built for mice.} On CalMS21, TREBA's released features and LISBET trail the recipe by 17.7 and 4.9
points, while HiDRA leads it by 3.5. That strength does not carry over. HiDRA reads pairs of
mice, so it cannot run on the flies or the human benchmarks. On the triplets, which are also in its pretraining data and
which it reads pair by pair (each mouse averaged over its two pairs), it trails the recipe by 7.3 points.

\paragraph{Fewer labels.} From about a tenth of the labels (8 of CalMS21's 70 training videos, and the same share
elsewhere), the recipe leads hBehaveMAE on all four benchmarks where we varied the labels (Figure~\ref{fig8:labels}).
With fewer labels, hBehaveMAE leads on the triplets, on Shot7M2 at the two smallest shares and on CalMS21 with two
labeled videos, where LISBET, CEBRA-Time, BAMS and body-frame hBehaveMAE also lead the recipe; HiDRA leads at every amount. Each method's
configuration is the one selected with all training labels, so the curves show how much the readout needs labels once
that configuration is fixed. Re-selecting the recipe's budget at each label amount on CalMS21 leaves these readings
unchanged (Appendix~\ref{app:fewlabels}). Where labels are scarcest, pretraining can pay off; from about a tenth of the
labels on, it does not beat the recipe here.

\paragraph{Learned readouts.} A per-frame linear classifier cannot integrate over time, while the window statistics do so
before any classifier sees them. We therefore also read every representation with a fixed one-hidden-layer MLP probe,
and with a temporal convolutional head whose context, training length and learning rate are chosen for each
representation on its own training data (Appendix~\ref{app:tunedhead}; Table~\ref{tab8:robust}). The head is not run on
the triplets, whose labels describe whole clips. On CalMS21 the temporal head lifts every representation, most of all
those the linear classifier reads worst, and brings half of them to 83 to 85: raw keypoints, the two variants, BAMS, LISBET and MARS's
features. hBehaveMAE stays behind (80.0) and HiDRA ahead (88.1). The recipe's lead over hBehaveMAE
also holds on hBABEL. On Shot7M2, hBehaveMAE leads the recipe and
is not resolved from the hand-written variant. Under the MLP probe the recipe leads every CalMS21 encoder but HiDRA.

\paragraph{Three levels of design.} Raw body-frame keypoints and velocities without windows, the generic recipe and the
hand-written variant form a ladder of hand design (Table~\ref{tab:raw}). Both variants lead hBehaveMAE in every column
of that table except Shot7M2 under the temporal head. Raw keypoints are the instructive case. Under the linear
classifier they trail hBehaveMAE on every benchmark, since a per-frame linear readout cannot integrate over time. Under
the temporal head, which does that integration, they lead hBehaveMAE on all three benchmarks (by 5.0, 0.5 and 6.3
points on CalMS21, Shot7M2 and hBABEL, each interval above zero) and score at least as high as the windowed recipe.
Read by a temporal model, then, hBehaveMAE's embeddings trail plain keypoints, and the window statistics give the linear
classifier much of what a temporal model would learn.

\begin{table}[t]
\centering\small
\caption{Raw keypoints (no windows), both variants and hBehaveMAE, all labels, under the linear classifier and the temporal head. $^\dagger$With the relations block; under the temporal head, raw rows are keypoints and velocities only.}\label{tab:raw}
\setlength{\tabcolsep}{3pt}
\fitbox{\linewidth}{%
\begin{tabular}{@{}lccccccc@{}}
\toprule
 & \multicolumn{2}{c}{CalMS21} & Triplets & \multicolumn{2}{c}{Shot7M2} & \multicolumn{2}{c}{hBABEL} \\
\cmidrule(lr){2-3}\cmidrule(lr){4-4}\cmidrule(lr){5-6}\cmidrule(lr){7-8}
 & Linear & Temporal & Linear & Linear & Temporal & Linear & Temporal \\
\midrule
Raw keypoints & 45.9$^\dagger$ & \textbf{85.0} & 68.7$^\dagger$ & 60.4 & \textbf{83.1} & 13.8 & \textbf{23.0} \\
Generic recipe & 79.7 & 83.4 & 83.1 & 66.0 & 79.4 & 20.9 & 22.7 \\
Hand-written variant & \textbf{80.1} & 83.9 & \textbf{83.7} & \textbf{68.8} & 82.4 & \textbf{22.1} & 20.9 \\
hBehaveMAE & 75.9 & 80.0 & 80.3 & 64.7 & 82.6 & 16.5 & 16.7 \\
\bottomrule
\end{tabular}}
\end{table}

\paragraph{Random weights.} A temporal head can do much of an encoder's work by itself (Table~\ref{tab:random}). This
control reads every arm with one fixed temporal head ($\pm$1\,s of context, 3,000 steps). With the same architecture and
input but untrained weights, BAMS read by this head is not resolved from trained BAMS on CalMS21 (four trained and five
random seeds, resampled with the videos: \gap{-0.6}{-1.4}{+0.5}), nor from the recipe, and it comes within 2.5 points of
trained BAMS on Shot7M2 and hBABEL. Under the linear classifier pretraining does help BAMS. Pretraining matters more for
hBehaveMAE on CalMS21, where it adds 13.9 to 16.2 points, than on Shot7M2, where it adds 2.7 to 4.3 and random
hBehaveMAE read by the fixed head is not resolved from the recipe under the same head (74.4 against 74.5). These encoder-matched
controls isolate what pretraining adds; comparisons with the recipe also differ in input and pipeline.

\begin{table}[!htb]
\centering\small
\caption{Random-weight controls, all labels: the same architecture and input with untrained weights against the trained encoder (ranges over seeds where several were run). Fixed temporal head: $\pm$1\,s, 3,000 steps. hBehaveMAE's arms are embedded every 10th frame and averaged over seeds, so its trained scores differ slightly from Tables~\ref{tab8:main} and~\ref{tab:raw}.}\label{tab:random}
\setlength{\tabcolsep}{3pt}
\fitbox{\linewidth}{%
\begin{tabular}{@{}llcc@{}}
\toprule
Encoder, benchmark & Readout & Random & Trained \\
\midrule
BAMS, CalMS21 & Linear & 45.9--49.8 & 55.2--56.6 \\
 & Fixed temporal head & 83.5--84.3 & 82.9--83.4 \\
BAMS, Shot7M2 & Linear & 46.8--48.2 & 57.3 \\
 & Fixed temporal head & 67.9--69.2 & 70.4 \\
BAMS, hBABEL & Linear & 12.0--13.1 & 14.5 \\
 & Fixed temporal head & 15.2--15.7 & 15.8 \\
hBehaveMAE, CalMS21 & Linear & 58.6 & 74.8 \\
 & Fixed temporal head & 65.8 & 79.7 \\
hBehaveMAE, Shot7M2 & Linear & 60.1 & 64.4 \\
 & Fixed temporal head & 74.4 & 77.0 \\
\bottomrule
\end{tabular}}
\end{table}

\paragraph{New annotators and new behaviors.} CalMS21's Tasks 2 and 3 reuse the Task 1 representations
(Appendix~\ref{app:tasks23}). With a new annotator the recipe leads hBehaveMAE by 3.4 points
and BAMS by 24.6; on new behaviors learned from a few videos neither gap is resolved, and HiDRA leads both tasks.

\section{Why a simple summary works}\label{sec8:needs}

\begin{figure*}[t]
\centering
\includegraphics[width=\textwidth]{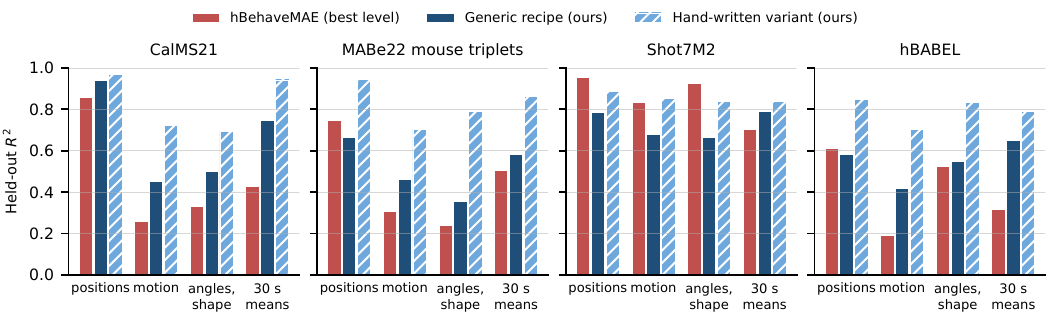}
\caption{How much of each kind of hand-written measurement a linear readout recovers from each frozen representation
(held-out $R^2$), for hBehaveMAE (its best level for each bar), the generic recipe and the hand-written variant on each
benchmark, on the same frames.}\label{fig8:encode}
\end{figure*}

\begin{table}[!htb]
\centering\small
\caption{Which windows carry the result, for both variants, with all labels. Encoder: hBehaveMAE; on the flies, BAMS's
published score. Sub-scale columns are in Appendix~\ref{app:carries}.}\label{tab8:windows}
\setlength{\tabcolsep}{3pt}
\fitbox{\linewidth}{%
\begin{tabular}{@{}lccccc@{}}
\toprule
Windows kept & CalMS21 & Triplets & Flies & Shot7M2 & hBABEL \\
\midrule
\multicolumn{6}{@{}l}{\emph{Generic recipe}} \\
Frame value only & 45.9 & 68.7 & 49.3 & 60.4 & 13.8 \\
Frame, 0.5\,s, 2\,s & \textbf{82.0} & 71.1 & \textbf{52.6} & 62.7 & 17.6 \\
8\,s, 30\,s, whole clip & 66.7 & 82.9 & 49.8 & 59.7 & 19.9 \\
All (the recipe) & 79.7 & 83.1 & 52.4 & 66.0 & 20.9 \\
\midrule
\multicolumn{6}{@{}l}{\emph{Hand-written variant}} \\
Frame value only & 68.3 & 70.2 & 45.9 & 60.0 & 15.2 \\
Frame, 0.5\,s, 2\,s & 80.7 & 72.7 & 47.5 & 64.2 & 17.5 \\
8\,s, 30\,s, whole clip & 65.6 & \textbf{84.7} & 45.5 & 59.9 & 20.9 \\
All & 80.1 & 83.7 & 51.6 & \textbf{68.8} & \textbf{22.1} \\
\midrule
Encoder & 75.9 & 80.3 & 48.2 & 64.7 & 16.5 \\
\bottomrule
\end{tabular}}
\end{table}

\paragraph{The body's own frame.} In the body's own frame a behavior looks the same wherever the animal is and
whichever way it faces, so a linear classifier need not learn it again for every position and heading. In camera, arena
or dish coordinates the same keypoints and velocities do much worse on every animal benchmark (Figure~\ref{fig8:ladder},
top). The largest effect is on CalMS21, where the body frame adds 20.7 points (\gap{+20.7}{+16.3}{+27.0}). Body-frame
input also lifts CEBRA-Time by 20 points and TS2Vec by 6 on CalMS21, while it leaves hBehaveMAE unchanged within its
interval; all stay below the recipe. Together with BAMS, which gets body-frame keypoints on every benchmark, and the
relational measurements supplied to hBehaveMAE in Section~\ref{sec8:why}, these are our matched-input checks: giving the
encoders the recipe's frame or its relational information narrows some gaps, most on CalMS21 with the
relational measurements, but leaves every encoder's score below the recipe's.

\paragraph{Windows of seconds.} The windows that matter follow the time scale of the labels (Table~\ref{tab8:windows}).
CalMS21's bouts last 1 to 2\,s, and the frame value with the 0.5 and 2\,s windows alone carries them (82.0). Long
windows carry the triplets' clip labels, Shot7M2's activities and hBABEL's sequence-level actions. Without windows the
recipe falls below the encoder on four benchmarks, and only the full set beats it on all five. The statistics matter as
well: MiniRocket and catch22 on the same body-frame keypoints trail the recipe on the flies, Shot7M2 and hBABEL
(Appendix~\ref{app:carries}).

\begin{table}[!htb]
\centering\small
\caption{Relational against individual hand-written measurements, each through the benchmark's full pipeline, all
labels. Encoder: hBehaveMAE; on the flies, BAMS's published score.}\label{tab8:rel}
\setlength{\tabcolsep}{3.5pt}
\fitbox{\linewidth}{%
\begin{tabular}{@{}lcccc@{}}
\toprule
Benchmark & Relational & Individual & All & Encoder \\
\midrule
CalMS21 & 79.9 & 64.2 & \textbf{80.1} & 75.9 \\
Mouse triplets & 68.6 & \textbf{83.9} & 83.7 & 80.3 \\
Fly groups & 43.8 & 47.5 & \textbf{51.6} & 48.2 \\
\bottomrule
\end{tabular}}
\end{table}

\paragraph{Relations where the labels are social.} In the recipe's per-mouse layout the relations block is its only information
about the other mouse: without it each mouse's own keypoints and velocities score 67.9 (the block adds
11.9 points), and the 11 relational hand-written measurements alone score 79.9 against 64.2 for the 12
individual ones (Table~\ref{tab8:rel}). Where the labels are not social, on the triplets and the flies, the block's effect is not resolved, and the individual measurements carry the triplets.

\section{What the encoders' embeddings make available}\label{sec8:why}

The analyses in this section are diagnostics: they describe what a readout can recover from frozen embeddings and how
that relates to scores, and they do not establish why an encoder scores as it does.

\begin{table}[!htb]
\centering\small
\caption{hBehaveMAE plain and through our window statistics, reduction and selection, with all labels (at its best
level; retrained on CalMS21, released elsewhere), against both of our variants.}\label{tab8:diag}
\setlength{\tabcolsep}{3pt}
\fitbox{\linewidth}{%
\begin{tabular}{@{}lcccc@{}}
\toprule
Benchmark & Plain & + our windows & Generic & Hand-written \\
\midrule
CalMS21 & 75.9 & 73.8 & 79.7 & \textbf{80.1} \\
Mouse triplets & 80.3 & 83.1 & 83.1 & \textbf{83.7} \\
Shot7M2 & 64.7 & 62.3 & 66.0 & \textbf{68.8} \\
hBABEL & 16.5 & 18.5 & 20.9 & \textbf{22.1} \\
\bottomrule
\end{tabular}}
\end{table}

\paragraph{Windows alone do not close the gap.} Given exactly our window statistics, reduction and selection,
hBehaveMAE still trails the recipe on CalMS21, Shot7M2 and hBABEL (Table~\ref{tab8:diag}), on CalMS21 even at 512
numbers per frame. Only on the triplets, whose labels describe whole clips, do the windows bring it level.

\paragraph{Less motion than the recipe makes available.} A ridge regression that reads each hand-written measurement
out of a frozen representation measures how much of it a linear classifier can use (held-out $R^2$). Every
representation yields positions more easily than motion, so we compare against a reference built from the same
keypoints: the recipe. It contains the keypoints' velocities, so it is a partial positive control; the hand-written
variant, built from these very quantities, is a full one. On the three real datasets, hBehaveMAE's embeddings yield
positions about as well as the recipe, but motion only about half as well and the 30\,s means less well too
(Figure~\ref{fig8:encode}). On the simulated Shot7M2 it is the reverse: hBehaveMAE carries more motion than the recipe.
A one-hidden-layer MLP reads little more motion out of any encoder, so to these probes the motion is less recoverable,
not merely encoded nonlinearly (Appendix~\ref{app:readouts}). Supplying the measurements' frame values to hBehaveMAE at
the same budget barely helps: motion adds at most 0.3 points, and only the mostly relational measurements help, on
CalMS21 (Table~\ref{tab:supply_bm}; Appendix~\ref{app:supply}).

\begin{table}[t]
\centering\small
\caption{The measurements' frame values supplied to hBehaveMAE at the budget, in place of part of its embedding: score, and gap to hBehaveMAE alone with its 95\% interval.}\label{tab:supply_bm}
\setlength{\tabcolsep}{3pt}
\fitbox{\linewidth}{%
\begin{tabular}{@{}lccc@{}}
\toprule
Benchmark & hBehaveMAE & + motion, at budget & + other, at budget \\
\midrule
CalMS21 & 75.9 & 76.1 ($+0.2$ [$-0.2$, $+0.6$]) & 78.8 ($+2.9$ [$+2.0$, $+3.6$]) \\
Mouse triplets & 80.3 & 79.5 ($-0.8$ [$-1.3$, $-0.4$]) & 78.7 ($-1.6$ [$-2.3$, $-0.9$]) \\
Shot7M2 & 64.7 & 64.9 ($+0.3$ [$+0.1$, $+0.4$]) & 64.3 ($-0.3$ [$-0.5$, $-0.2$]) \\
hBABEL & 16.5 & 16.7 ($+0.1$ [$-0.2$, $+0.4$]) & 16.9 ($+0.4$ [$-0.1$, $+0.8$]) \\
\bottomrule
\end{tabular}}
\end{table}

\begin{figure}[t]
\centering
\includegraphics[width=\linewidth]{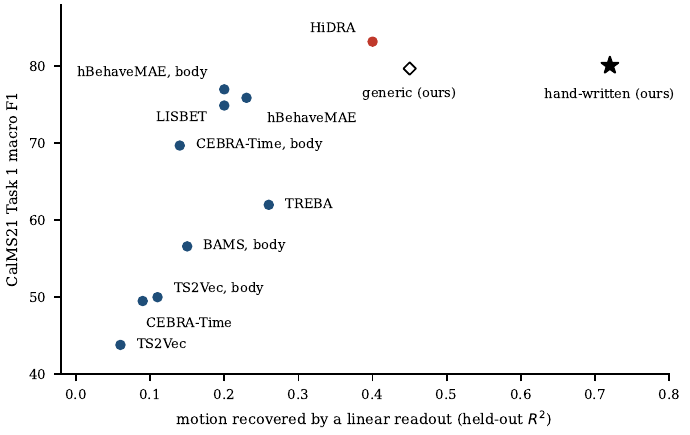}
\caption{CalMS21: motion recovered by a linear readout against Task 1 score for ten keypoint-trained representations (HiDRA in red). Our two variants are shown for reference and are not part of the correlation.}\label{fig8:scatter}
\end{figure}

\paragraph{Motion tracks the ranking.} Across ten keypoint-trained representations on CalMS21, those from which a
linear readout recovers more motion score higher (Spearman 0.82, exact permutation $p = 0.006$, two-sided;
Figure~\ref{fig8:scatter}). Those that recover more of the 30\,s means (0.78) and of the angles and shape (0.65) also
score higher, so this test does not single out motion. Three encoders appear with two inputs, so the ten points are not
independent; with one point per encoder the correlation is 0.86, and controlling for the input frame it is 0.80
(Appendix~\ref{app:readouts}). With seven distinct encoders that differ in architecture and input, this is weak,
correlational evidence, not a test of cause. HiDRA, trained to forecast where the keypoints of both mice go rather than
to reconstruct where they are, carries the most motion of any encoder trained on keypoints and is the one encoder that
beats the recipe on CalMS21. Motion is not the whole story, though: the recipe carries more motion than HiDRA, which
carries more of the angles and shape and learns the relations between the mice. Encoders that carry motion over several
time scales, as HiDRA's forecasting objective begins to, are a natural next step.

\section{Limitations}\label{sec8:limits}

We compare frozen representations under the benchmarks' linear protocol. Learned readouts narrow some gaps
(Section~\ref{sec8:results}), and fine-tuned or end-to-end models, some of which score higher on
CalMS21~\citep{sun2021calms21}, may compare differently. Outside CalMS21 the encoders are mostly the released
hBehaveMAE and BAMS retrained by us; on the flies BAMS enters through its published score, without a paired interval,
and MICE~\citep{su2026mice} enters only through its published numbers, as its linked repository held no code or
weights as of November 2026. The triplets and flies test clip- and group-level traits rather than behaviors; on the
flies the score comes mostly from fly type and line category, and the recipe is near the random sample submission on
stimulation control and the per-fly behaviors (Table~\ref{tab:fly_categories}). We use each benchmark's official splits. CalMS21's tasks come from one
laboratory, and its 19 test videos make intervals wide. Where motion is finer than the recipe's summaries, as for Shot7M2's movemes, hBehaveMAE keeps an edge.

Two settings change the reading. On Shot7M2, learned readouts put hBehaveMAE ahead of the generic recipe
(Table~\ref{tab8:robust}). With the fewest labels, pretrained encoders can lead (Section~\ref{sec8:results}).

\section{Conclusion}

Across mice, flies and humans, encoders pretrained on each benchmark's own data rarely beat a training-free summary of
body-frame keypoints over windows from half a second to the whole clip, and plain measurements in their place add a
resolved gain on the two human-skeleton benchmarks. The readouts suggest why
the encoders fall short, without proving it: on the real datasets their embeddings make about as much of where body parts
are recoverable as the recipe does, but about half as much of how they move. We hypothesize that objectives carrying motion over several time scales, tested on benchmarks whose labels are
behaviors at the scale of bouts, are the clearest way forward. The recipe runs in minutes on a CPU, and on benchmarks like these it is a natural reference to report alongside new
behavior encoders.

\paragraph{Code and data.} \codestatement{} The datasets are available from their original sources.

\paragraph{Use of AI assistants.} \aistatement{}

\section*{Broader impact}

This paper aims to improve how representations of animal and human behavior are evaluated. A training-free baseline
that runs in minutes on a CPU lowers the computational cost of behavior analysis and gives laboratories without large
compute budgets a strong reference point; reporting it next to new encoders should make claimed gains easier to
interpret. The benchmarks we use are public, and the human data are anonymized motion capture. Behavior analysis
supports neuroscience and animal-welfare research and could also be applied to monitoring people; our work does not add
capabilities beyond those of the existing benchmarks and encoders, and we see no consequences that must be highlighted
here.

\bibliographystyle{plainnat}
\bibliography{refs}

\appendix
\setcounter{totalnumber}{6}\setcounter{topnumber}{4}\setcounter{bottomnumber}{3}
\renewcommand{\topfraction}{0.95}\renewcommand{\bottomfraction}{0.9}\renewcommand{\textfraction}{0.05}\renewcommand{\floatpagefraction}{0.85}

\section*{Appendices}

Appendix~\ref{app:recipe} gives the recipe's input on each benchmark and the hand-written measurements;
Appendix~\ref{app:protocol} the protocol, the encoders' training and the hBABEL files;
Appendix~\ref{app:benchmarks} each benchmark's own scores; Appendices~\ref{app:robust} to~\ref{app:diag} further checks
and analyses; and Appendix~\ref{app:further} the further encoders run on CalMS21. Unless a table says otherwise, the checks cover both variants; the cuts to 32 and 64
numbers per frame and the tests of Table~\ref{tab:why} use the hand-written variant. Gaps are in points of each
benchmark's own score, with paired 95\% intervals from 10,000 resamples of the test videos, clips, episodes or
sequences, computed from unrounded scores.

\FloatBarrier
\section{The recipe and the hand-written variant}\label{app:recipe}

\subsection{The generic recipe's input}

Table~\ref{tab:generic_input} gives each benchmark's input and the layout its cross-validation chose.

The
criterion is each benchmark script's own cross-validated score on training data (CalMS21: five-fold macro F1 over the 70
training videos; the others: the mean F1 of the script's development tasks), the same rule our hand-written variant
and every encoder variant followed.

\begin{table}[!htbp]
\centering\small
\caption{The generic recipe's input on each benchmark, and the layout its cross-validation chose on training data (A:
one PCA of means and spreads; B: A with histograms; C: two PCAs, short and long windows; D: C with histograms).}\label{tab:generic_input}
\setlength{\tabcolsep}{3pt}
\fitbox{\linewidth}{%
\begin{tabular}{@{}lllllc@{}}
\toprule
Benchmark & Points per body & Body frame (origin; heading) & Units; smoothing & Animals pooled by & Numbers \\
\midrule
CalMS21 & 7 keypoints, 2 mice & mean of neck and hips; to the nose & body lengths; 3-frame median & roles kept (resident first) & 70 \\
MABe22 triplets & 12 keypoints, 3 mice & center keypoint; tail base to neck & 100 px; gaps interpolated & sorted per number & 165 \\
MABe22 flies & 20 landmarks, $\le$11 flies & center; abdomen tip to head & mm; 5-frame median (150 Hz) & mean, spread, min, max & 348 \\
Shot7M2 & 26 joints (3D) & pelvis; hip line & player size; none & one body & 156 \\
hBABEL & 25 joints (3D) & pelvis; hip line & body size; none & one body & 150 \\
\bottomrule
\end{tabular}}

\vspace{4pt}
\fitbox{\linewidth}{%
\begin{tabular}{@{}ll@{}}
\toprule
Benchmark & Layout criterion on training data (chosen in bold) \\
\midrule

CalMS21 & A 0.640, \textbf{B 0.760}, C 0.633, D 0.744 \\
MABe22 triplets & A 0.390, B 0.380, \textbf{C 0.396}, D 0.388 \\
MABe22 flies & A 0.579, B 0.651, C 0.619, \textbf{D 0.709} \\
Shot7M2 & A 0.561, B 0.521, \textbf{C 0.563}, D 0.522 \\
hBABEL & A 0.213, \textbf{B 0.220}, C 0.215, D 0.213 \\
\bottomrule
\end{tabular}}
\end{table}

Velocities are per second (central differences; over $\pm$5 frames for the flies). Each animal benchmark adds the relations
block (Section~\ref{sec8:method}) to every animal's numbers before pooling. The window statistics, standardization, PCA
and budget are those of the hand-written variant on the same benchmark.

\subsection{The hand-written measurements}

For CalMS21 (two mice, seven keypoints each, 30\,Hz) the hand-written variant uses 23 measurements
(Table~\ref{tab:measurements}). They describe generic relations between two agents, not definitions of behaviors. Keypoints are first smoothed with a three-frame median filter. Distances are in body
lengths, the clip's median nose-to-tail distance, and speeds in body lengths per second, so camera scale drops out. No
measurement uses the arena's position or axes. The windows are those of the generic recipe; cross-validation chose two
PCAs with histograms: 64 numbers from the frame value and the two short windows, and 64 from the two long windows.

\begin{table}[!htbp]
\centering\small
\caption{The 23 CalMS21 measurements, in seven groups.}\label{tab:measurements}
\fitbox{\linewidth}{%
\begin{tabular}{@{}lp{0.78\linewidth}@{}}
\toprule
Group & Measurements \\
\midrule
Distance & distance between the two body centers; one minus that distance; contact (below one body length) \\
Speeds & each mouse's center speed \\
Approach & closing speed; each mouse's speed toward the other \\
Activity & per mouse, the share of still frames (below 0.25 body lengths per second) and the rate of switching between
still and moving, over 0.5\,s; the correlation of the two speeds over 0.5\,s \\
Nose & each mouse's nose to the nearest keypoint of the other \\
Heading & whether each mouse's head points at the other (cosine); how fast each head turns \\
Posture & each mouse's body length relative to its median; nose speed minus body speed \\
\bottomrule
\end{tabular}}
\end{table}

\paragraph{MABe22 mouse triplets.} The variant uses 89 measurements written for three mice: per-mouse speeds, body
lengths, turning and position in the arena, plus pairwise distances, closing speeds, facing and approach. Per-mouse
and pairwise values are sorted across animals, so the representation does not depend on mouse order. The whole-clip
window spans the 60-second clip. Everything is fitted on the
1,600 user-train clips, and cross-validation there chose two PCAs without histograms.

\paragraph{MABe22 fly groups.} For groups of up to 11 flies at 150\,Hz we wrote 24 generic measurements per fly
(speeds, turning, body and wing posture, leg reach, and relations to the nearest fly), pooled over flies by mean,
spread, minimum and maximum, plus four group measurements: 100 numbers per frame. Windows run from 0.5\,s to the whole
30\,s clip, and PCA reduces to the benchmark's limit of 256. Everything is fitted on the 426 user-train clips, and
cross-validation there chose one PCA of means and spreads.

\paragraph{Shot7M2.} For one simulated 3D body (26 joints, 30\,Hz) we wrote 40 measurements: body size; pelvis and
head height; the pelvis's vertical speed and the body's and the head's speed; for each wrist and ankle, its position in
the body's frame (three numbers), its speed and its vertical speed; elbow and knee angles and their rates; trunk lean
and twist; signed turning speed; hand and foot gaps; and stillness. Left and
right limbs are kept apart. The windows are those of the MABe22 mice, PCA reduces to the benchmark's 64 numbers, and
cross-validation on the 1,280 user-train episodes chose two PCAs of means and spreads.

\paragraph{hBABEL.} For real human motion capture we use the same 40 measurements on the 25-joint skeleton that
hBehaveMAE receives (BehaveMAE's processed joints: each frame's joint centroid is fixed, so there is no travel through
the room). Heights are therefore measured from the lower ankle in the same frame. Everything else is as for Shot7M2,
including the 64 numbers per frame (the evaluator's limit): the statistics are fitted on BABEL's 6,601 training
sequences without labels, and cross-validation on the training
sequences' own labels chose the same layout.

\paragraph{How much is hand-built.} Each measurement set is 23 to 100 ordinary kinematic quantities computed from
the tracked points: positions, speeds, distances, angles and heights, of the kind found in any behavior-analysis
toolkit. None of them defines a behavior, and none uses a label. Each dataset's per-frame measurement function is
under 100 lines of Python (46 for CalMS21, 82 for the MABe22 mice, 78 for the flies, 43 for Shot7M2, 44 for hBABEL). Fitting the
window statistics and PCA takes minutes on a CPU; training BAMS on CalMS21 alone took us 5 to 6 hours on a GPU.
The last row of Table~\ref{tab8:main} measures how much they add over the generic recipe. All
runs used one workstation with an Intel Core i9-12900KF CPU (16 cores), 64 GB of RAM and an NVIDIA GeForce RTX 4080
SUPER GPU (16 GB).

\FloatBarrier
\section{Protocol details}\label{app:protocol}

\paragraph{One classifier, frozen representations.} No method gives more numbers per frame than the benchmark's budget
(LISBET's and TREBA's released features give 32 on CalMS21, against 128), and the encoders stay frozen. Wider encoder
outputs are reduced by their authors' own procedure: hBehaveMAE's script reduces its wider levels to 128 with a PCA
fitted on the frames it embeds, which include the test videos and likely helps it. The ridge classifier is fitted on
the frames of the training clips. Where a method comes in several variants (hBehaveMAE's three levels, our four
layouts, BAMS's embedding shifted 0.5 or 1\,s later, since its embedding of a frame sees only the past), one variant is
chosen by five-fold cross-validation over the training videos with the same classifier. Output size and pretraining
data are matched; model size follows each author's recipe.

\paragraph{Label budgets.} On CalMS21 the classifier is also trained on 2, 4, 8, 16 or 35 labeled training videos, ten
fixed sets per size, each containing all three behaviors (only 5 of the 70 videos contain all three, so ten distinct
sets need at least two videos), and the scores of the ten sets are averaged. On the mouse triplets, Shot7M2 and hBABEL, whose evaluators train their classifier on part of
the benchmark's test-time clips (2,668 triplet clips, 2,040 Shot7M2 episodes, 869 hBABEL sequences) and score the rest
(1,068, 680 and 290), apart from the training data our statistics are fitted on, the same budgets are shares of each
evaluator's training list (2, 4, 8, 16 and 35 of 70), with five random subsets each
(Figure~\ref{fig8:labels}). Each method keeps
the variant it chose with all labels. Everything else on CalMS21 Task 1 uses all labels.

\paragraph{Encoders.} BAMS and hBehaveMAE were retrained on CalMS21 with their authors' code and recipes on its
unlabeled and training videos: hBehaveMAE for 200 epochs, BAMS for 500. BAMS takes both mice's keypoints and their
velocities in the resident's body frame and, as in its authors' generic keypoint template, predicts the velocities.
Its authors' data loader divides each input by its largest absolute value; in camera coordinates rare tracking jumps
(a median of 709 pixels per input, up to 989) would set that value and shrink the typical motion of about 2 pixels per frame to about 0.003, so we
scale its input robustly (median and interquartile range, clipped) and no jump sets the scale. hBehaveMAE is read at each of its three levels (0, 1 and 2, from the shortest time
scale to the longest), and the best on training videos is used; it was also trained on body-frame keypoints, while its
entry keeps its authors' input. On MABe22 we used the authors' released hBehaveMAE checkpoint and the benchmark's own
evaluator, which fits the ridge classifier on three random splits of its training list (three seeds) and combines
their predictions by majority vote. On the fly groups we used the MABe 2022 public evaluator, which reproduces the 50
published scores of the benchmark's random sample submission. On Shot7M2 and hBABEL we used the released hBehaveMAE
checkpoints and the authors' evaluator, and retrained BAMS with its authors' code on each benchmark's training data for
500 epochs, given each body's joints in its own heading frame with their velocities (robust-scaled) and with the channel
widths of its mouse-triplet recipe, which give the benchmarks' 64 numbers per frame. For the released hBehaveMAE on
those benchmarks we report its best level on the test set, a choice that favors it; its embeddings come at each
benchmark's budget through its authors' own PCA.

\paragraph{Reproducing the opponents.} Read with a one-hidden-layer MLP probe, as
MICE~\citep{su2026mice} reads it, our hBehaveMAE scores 82.4 on CalMS21, above the 76.05 that MICE reports for it (75.9
with the linear classifier), so its retraining did not handicap it. Our BAMS scores 56.6 with the linear classifier and 72.7 when read with a
one-hidden-layer MLP probe (Table~\ref{tab:readouts}), as MICE's numbers are, close to the 74.98 that MICE reports for
BAMS on CalMS21. Our TS2Vec runs land within
about 3 points of the TS2Vec scores the hBehaveMAE paper reports (59.6 on Shot7M2, 14.4 on hBABEL), so they are not
unusually weak. CalMS21's poses come from MARS, and Task 1's videos match the training and test sets of MARS's
classifiers, which use hand-designed features; the benchmark paper~\citep{sun2021calms21} reports 79.3 for a supervised 1D
convolutional network, 82.9 when that network is given TREBA's self-supervised features, and 86.4 for the 2021
challenge winner, a semi-supervised model trained with the labels, where our scores come from frozen representations
and a linear classifier.

\paragraph{MICE.} The repository linked from MICE's paper (\url{https://github.com/eric0708/MICE}) contained no code
or weights as of November 2026, so we report its published numbers. Its reported CalMS21 score, 83.3 (pretrained on CalMS21 alone, individual
and social embeddings concatenated; its Tables~1 and~3), comes from what it calls linear probing,
a one-hidden-layer MLP on frozen features, so it cannot be compared with our numbers under the ridge classifier; under
our own fixed one-hidden-layer MLP probe the generic recipe scores 85.8 (Table~\ref{tab:readouts}), though the two
probes differ.

\paragraph{Uncertainty.} To cover training luck, hBehaveMAE, TS2Vec and CEBRA-Time were trained three times on CalMS21 with
different seeds and BAMS four times (Appendix~\ref{app:robust}); elsewhere every encoder was trained or run once. On CalMS21, attack occurs in only 5 of
the 19 test videos, so resampling videos moves its share a lot and skews some intervals; resamples without any attack
video (0.3\%) score its F1 as 0. Every gap is read by its percentile interval; bias-corrected intervals and paired
permutation tests are in Appendix~\ref{app:robust}.

\paragraph{Generic time-series features.} MiniRocket (10,000 kernels) and
catch22 with each series' mean and spread (catch24), from aeon 1.2.0~\citep{middlehurst2024aeon} at its default settings,
read each body's keypoints and velocities in its own frame, at 30 frames per second, reduced to 16 channels by a PCA
on training frames; windows of 2\,s and 8\,s centered every 0.5\,s (each frame takes the nearest) and the whole clip
resampled to 300 steps; each window length's features are standardized and reduced by PCA to a third of the budget,
all fitted on training windows only.

\subsection{What each method was given}\label{app:given}

On CalMS21 Task 1, every encoder learned from the same unlabeled videos as far as its release allows, and every method
fed the same classifier at most 128 numbers per frame (Table~\ref{tab:given}). Neither variant has trained parameters; what is
fitted on unlabeled and training videos are the histogram edges, the moments and the PCA (for the hand-written
variant on CalMS21, 161 bin edges,
943 means and 943 standard deviations, and 57,792 PCA loadings of two PCAs from 463 and 440 standardized statistics to
64 numbers each: 59,839 numbers in all; the other 40 statistics are histogram bins that no training frame falls in, and are dropped). Parameter counts are read
from the checkpoints we scored.

\begin{table}[!htbp]
\centering\small
\caption{What each method was given on CalMS21 Task 1: trained parameters, numbers per frame and
the unlabeled videos it learned from.}\label{tab:given}
\fitbox{\linewidth}{%
\begin{tabular}{@{}lrcl@{}}
\toprule
Method & Trained parameters & Numbers per frame & Unlabeled videos it learned from \\
\midrule
Generic recipe (ours) & 0 (fitted statistics only) & 128 & CalMS21 unlabeled and training \\
Hand-written variant (ours) & 0 (59,839 fitted statistics) & 128 & CalMS21 unlabeled and training \\
MARS's hand-built features & 0 (fitted PCA only) & 128 & CalMS21 training (PCA only) \\
BAMS & 1,397,266 & 128 & CalMS21 unlabeled and training \\
hBehaveMAE (encoder) & 19,856,448 & 128 & CalMS21 unlabeled and training \\
hBehaveMAE, body-frame keypoints & 19,856,448 & 128 & CalMS21 unlabeled and training \\
TS2Vec & 331,201 & 128 & CalMS21 unlabeled and training \\
CEBRA-Time & 845,056 & 128 & CalMS21 unlabeled and training \\
LISBET, released & 58,208 & 32 & CalMS21 unlabeled \\
TREBA, released & not released (features only) & 32 & its authors' \\
HiDRA, released (five trunks) & 139,972,466 & 128 & 9,008 videos, CalMS21 among them \\
\bottomrule
\end{tabular}}
\end{table}

On the released benchmarks the encoders are the authors' released hBehaveMAE checkpoints (encoders of 20,073,408,
10,935,982 and 15,354,784 parameters for the mouse triplets, Shot7M2 and hBABEL), each pretrained on its benchmark's
own data; BAMS trained by us on Shot7M2 and hBABEL (2,151,282 and 2,095,122 parameters, 64 numbers per frame); and
TS2Vec and CEBRA-Time trained by us on each benchmark's training clips.

\subsection{Building hBABEL's files}\label{app:hbabel}

hBABEL's files are not distributed. We downloaded the 17 AMASS subsets that BABEL's labels refer to (SMPL+H~G),
TEACH's~\citep{athanasiou2022teach} processed BABEL labels and the SMPL+H body model, ran TEACH's AMASS processing and split scripts and then
BehaveMAE's two hBABEL scripts, all unmodified except for the split script's three file paths. This gives 13,282
processed sequences, of which 10,869 are in BABEL, and 1,159 validation sequences with labels, the test set. The
evaluator splits these into 869 sequences for training its classifiers and 290 for testing them; our statistics are
fitted without labels on BABEL's 6,601 training sequences.

The evaluator reads its own task list, split and frame map, and refuses a submission whose frame map differs. Our
files hold the same 1,159 sequences with the same lengths, but listed in another order, which changes the split of
those sequences into classifier-training and test, every frame offset, and one of the 180 tasks at the cut that keeps the 120 most frequent frame-level behaviors (the other 60 tasks are
the most frequent sequence-level behaviors).
We therefore rebuilt the label file in the evaluator's exact layout, with its 180 tasks and its split, using
BehaveMAE's labelling loop copied verbatim (checked first: on our own layout the copy reproduces BehaveMAE's label file
exactly), and moved every submission into that layout sequence by sequence. On the 179 shared tasks the rebuilt labels
differ from BehaveMAE's own file in 591 of about 103 million task-frames, all on the first frame of a sequence. The released hBehaveMAE, run with
its authors' test settings on these files, reproduces the score computed from its published columns approximately (16.95 against 16.98, taking each
column from its best level): its level 0 matches the published frame-level columns to the decimal, and its level 2 is
within 0.4 of the published sequence-level columns.

\FloatBarrier
\section{Per-benchmark results}\label{app:benchmarks}

Each benchmark's own scores for both variants, with the published rows its papers report, behind Table~\ref{tab8:main}.

\subsection{CalMS21}

Tables~\ref{tab:calms21} and~\ref{tab:perbehavior} give the scores with all labels and the gaps for each behavior. Against each of hBehaveMAE's three training seeds (75.9, 75.2 and 73.8) the generic recipe leads by 3.8 to 5.9 points and the hand-written variant by 4.2 to 6.3, every interval above zero.

\begin{table}[!htbp]
\centering\small
\caption{CalMS21 Task 1 with all labels, macro F1 on the 19 test videos, with 95\% intervals, for encoders trained
with seed 0, and hBehaveMAE averaged over three training seeds.}\label{tab:calms21}
\fitbox{\linewidth}{%
\begin{tabular}{@{}lc@{}}
\toprule
Method & Macro F1 \\
\midrule
Generic recipe (ours) & 79.7 [75.1, 84.0] \\
Hand-written variant (ours) & \textbf{80.1} [75.9, 83.3] \\
hBehaveMAE & 75.9 [70.6, 80.2] \\
BAMS & 56.6 [48.9, 59.8] \\
\midrule
hBehaveMAE, mean of 3 seeds & 75.0 \\
\bottomrule
\end{tabular}}
\end{table}

\begin{table}[!htbp]
\centering\small
\caption{CalMS21 Task 1 per behavior, all labels and seed-0 encoders: F1, and each variant minus each encoder with
paired 95\% intervals. The lead over hBehaveMAE comes from attack and investigation.}\label{tab:perbehavior}
\fitbox{\linewidth}{%
\begin{tabular}{@{}lccc@{}}
\toprule
 & Attack & Investigation & Mount \\
\midrule
Generic recipe (ours) & 71.4 & 78.3 & 89.5 \\
Hand-written variant (ours) & 70.0 & 80.5 & 89.9 \\
HiDRA, released & 71.5 & 85.4 & 92.8 \\
hBehaveMAE & 65.2 & 72.4 & 90.1 \\
LISBET, released & 60.6 & 77.2 & 86.7 \\
BAMS & 45.1 & 60.4 & 64.2 \\
\midrule
Generic minus hBehaveMAE & \gap{+6.3}{+3.1}{+16.3} & \gap{+5.9}{+3.2}{+9.0} & \gap{-0.7}{-1.7}{+0.5} \\
Generic minus BAMS & \gap{+26.3}{+15.3}{+54.5} & \gap{+18.0}{+14.6}{+20.9} & \gap{+25.3}{+16.6}{+40.9} \\
Generic minus HiDRA & \gap{0.0}{-1.0}{+5.1} & \gap{-7.0}{-9.7}{-5.1} & \gap{-3.3}{-5.4}{-2.1} \\
Hand-written minus hBehaveMAE & \gap{+4.8}{+0.0}{+10.5} & \gap{+8.1}{+5.7}{+11.0} & \gap{-0.3}{-1.3}{+0.8} \\
Hand-written minus BAMS & \gap{+24.9}{+14.6}{+48.1} & \gap{+20.2}{+16.1}{+24.0} & \gap{+25.7}{+17.1}{+41.1} \\
Hand-written minus HiDRA & \gap{-1.5}{-5.4}{+1.9} & \gap{-4.8}{-6.6}{-3.4} & \gap{-2.9}{-5.2}{-1.8} \\
\bottomrule
\end{tabular}}
\end{table}

\subsection{CalMS21 Tasks 2 and 3}\label{app:tasks23}

CalMS21's Tasks 2 and 3 reuse the Task 1 representations, frozen,
with the same ridge classifier (Table~\ref{tab:tasks23}). Task 2 trains on each of five further annotators' six
training videos and tests on that annotator's test videos; Task 3 learns each of seven new behaviors from its few
labeled videos. With a new annotator the generic recipe leads hBehaveMAE by 3.4 points (\gap{+3.4}{+1.4}{+5.6}) and BAMS by 24.6, and is not resolved from the hand-written variant. New behaviors learned from a few videos are harder
for every method: there the recipe is not resolved from hBehaveMAE (\gap{-2.5}{-6.5}{+0.2}) or BAMS, and trails the hand-written variant by 3.6. HiDRA leads in both settings, by
6.9 and 16.5 points.

\begin{table}[!htbp]
\centering\small
\caption{CalMS21 Tasks 2 and 3 with the Task 1 representations, frozen, and the shared ridge classifier (encoders of
seed 0). Task 2: macro F1 per new annotator, averaged over the five annotators. Task 3: F1 per new behavior, averaged
over the seven behaviors. Bottom rows: paired 95\% intervals. $^\ast$Pretrained on pooled data that include
CalMS21.}\label{tab:tasks23}
\setlength{\tabcolsep}{4pt}
\fitbox{\linewidth}{%
\begin{tabular}{@{}lcc@{}}
\toprule
 & Task 2: new annotators & Task 3: new behaviors \\
\midrule
Generic recipe (ours) & 72.5 & 24.4 \\
Hand-written variant (ours) & 74.0 & 27.9 \\
hBehaveMAE & 69.2 & 26.9 \\
BAMS & 47.9 & 23.1 \\
HiDRA, released$^\ast$ & \textbf{79.5} & \textbf{40.8} \\
\midrule
Generic minus hBehaveMAE & \gap{+3.4}{+1.4}{+5.6} & \gap{-2.5}{-6.5}{+0.2} \\
Generic minus BAMS & \gap{+24.6}{+17.6}{+27.5} & \gap{+1.3}{-1.9}{+3.5} \\
Generic minus HiDRA & \gap{-6.9}{-9.2}{-3.6} & \gap{-16.5}{-20.6}{-13.2} \\
Hand-written minus generic & \gap{+1.5}{-0.3}{+2.5} & \gap{+3.6}{+0.9}{+7.2} \\
\bottomrule
\end{tabular}}
\end{table}

\subsection{MABe22 mouse triplets}

\begin{table}[!htbp]
\centering\small
\caption{MABe22 mouse triplets, benchmark evaluator. Clip-level F1 is the mean of strain and lights; brackets are
95\% intervals from resampling test clips. The evaluator's 13 tasks are these 11, scored by F1, and two regressions (experiment day and time of day). Published rows are from the hBehaveMAE paper~\citep{stoffl2024hbehavemae},
Table~2 (``transductive'': pretrained on the test clips too).}\label{tab:mabe22}
\fitbox{\linewidth}{%
\begin{tabular}{@{}lcc@{}}
\toprule
Method & Clip-level F1 & Frame-level F1 \\
\midrule
Generic recipe (ours) & 83.1 [81.3, 84.7] & 21.8 \\
Hand-written variant (ours) & \textbf{83.7} [81.9, 85.3] & \textbf{24.5} \\
hBehaveMAE, released, best level (our run) & 80.3 [78.7, 81.8] & 22.3 \\
HiDRA, released, five trunks (our run) & 75.8 [74.1, 77.4] & 21.3 \\
\midrule
hBehaveMAE, published, transductive & 80.3 & 22.3 \\
BehaveMAE, published, transductive & 81.0 & 21.5 \\
BAMS, published, transductive & 80.1 & 19.8 \\
hBehaveMAE, published, inductive & 78.5 & 20.1 \\
\bottomrule
\end{tabular}}
\end{table}

The evaluator's 13 tasks are the two clip-level tasks behind the main score (mouse strain and whether the lights are
on), nine frame-level behaviors and two regressions. The frame-level behaviors, such as chases and huddles, were labeled
from the poses by the benchmark's own rules; on their mean F1 the hand-written variant scores 24.5, hBehaveMAE 22.3 and
the generic recipe 21.8. On the regressions, the
experiment day and the time of day, all three are level (mean squared error 0.093 and 0.092 for the generic recipe,
0.092 and 0.091 for the hand-written variant, 0.092 and 0.092 for hBehaveMAE).

\subsection{MABe22 fly groups}\label{app:flies}

Table~\ref{tab:flies} adds the published rows, and Table~\ref{tab:fly_categories} splits the score by task category.

\begin{table}[!htbp]
\centering\small
\caption{MABe22 fly groups, mean F1 over 50 tasks, benchmark scorer. Published rows are from Table~4 of the
NeurIPS 2023 version of the BAMS paper~\citep{azabou2023bams}.}\label{tab:flies}
\fitbox{\linewidth}{%
\begin{tabular}{@{}lc@{}}
\toprule
Method & Mean F1 \\
\midrule
Generic recipe (ours) & \textbf{52.4} \\
Hand-written variant (ours) & 51.6 \\
BAMS & 48.2 \\
T-GPT & 45.8 \\
T-Perceiver & 44.8 \\
PCA & 42.5 \\
TVAE & 37.0 \\
\bottomrule
\end{tabular}}
\end{table}

Generic minus the same without the block
\gap{-0.6}{-2.5}{+1.1}; minus dish coordinates \gap{+10.0}{+6.8}{+12.6}; the hand-written variant minus the
generic recipe \gap{-0.8}{-3.1}{+1.7}.

\begin{table}[!htbp]
\centering\small
\caption{MABe22 fly groups, mean F1 over the 50 tasks and over each task category (number of tasks in brackets), all
labels, on the benchmark's scorer: the generic recipe in the layout its cross-validation chose (D), the same without the
relations block and in dish coordinates; the hand-written variant pooled over flies (its layout) and with the flies kept
apart; and the random sample submission, which shows each category's chance
level, high for some (72.1 on female versus male).}\label{tab:fly_categories}
\setlength{\tabcolsep}{3pt}
\fitbox{\linewidth}{%
\begin{tabular}{@{}lccccccc@{}}
\toprule
Representation & All (50) & Fly type (11) & Stim.\ control (10) & Stim.\ aggression (15) & Line category (7) & Female vs male (1) & Manual, per fly (6) \\
\midrule
Generic recipe (ours), layout D & 52.4 & 57.3 & 47.9 & 52.1 & 74.2 & 87.3 & 20.4 \\
Without the relations block, layout D & 53.0 & 59.2 & 47.1 & 53.8 & 74.1 & 87.2 & 19.2 \\
Dish coordinates, layout D & 42.4 & 33.2 & 47.6 & 52.0 & 45.4 & 86.1 & 15.8 \\
Hand-written variant (ours), pooled & 51.6 & 50.6 & 52.8 & 55.0 & 68.7 & 88.0 & 17.0 \\
Hand-written variant, flies kept apart & 48.8 & 41.2 & 47.7 & 56.0 & 60.5 & 97.6 & 24.8 \\
\midrule
BAMS, published & 48.2 & & & & & & \\
Random sample submission & 35.7 & 14.5 & 48.4 & 49.6 & 28.4 & 72.1 & 21.0 \\
\bottomrule
\end{tabular}}
\end{table}

The score comes from group-level tasks such as genotype and line category. Both variants pool over flies, which suits
these tasks; keeping the flies apart lifts the six per-fly behaviors but costs most group tasks
(Table~\ref{tab:fly_categories}).

\subsection{Shot7M2}

\begin{table}[!htbp]
\centering\small
\caption{Shot7M2, the benchmark's evaluator. All F1 is the mean of the three scale scores. hBehaveMAE's scale columns
in its paper are each the best of its levels, whose mean is 68.9; its All F1 is one level. Published rows are from the hBehaveMAE
paper~\citep{stoffl2024hbehavemae}, Table~3.}\label{tab:shot}
\fitbox{\linewidth}{%
\begin{tabular}{@{}lcccc@{}}
\toprule
Method & All F1 & Activities & Actions & Movemes \\
\midrule
Generic recipe (ours) & 66.0 & 90.4 & 55.5 & 52.2 \\
Hand-written variant (ours) & \textbf{68.8} & \textbf{92.6} & 55.2 & 58.7 \\
hBehaveMAE, released, best level (our run) & 64.7 & 76.4 & 54.8 & 62.9 \\
hBehaveMAE, best level for each scale (our run) & 68.9 & 80.9 & 58.5 & 67.3 \\
BAMS, retrained (our run) & 57.3 & 73.7 & 45.4 & 52.8 \\
\midrule
hBehaveMAE, published & 64.7 & 80.9 & 58.5 & 67.3 \\
BehaveMAE, published & 64.3 & 76.7 & 55.1 & 61.0 \\
PCA-30, published & 61.9 & 68.3 & 50.8 & 66.7 \\
PCA, published & 60.8 & 69.5 & 51.0 & 61.8 \\
TS2Vec, published & 59.6 & 68.6 & 50.0 & 60.2 \\
BAMS, published & 53.7 & 67.1 & 45.3 & 48.6 \\
TVAE, published & 51.2 & 65.5 & 47.0 & 41.0 \\
\bottomrule
\end{tabular}}
\end{table}

The released hBehaveMAE reproduces its published scores (Table~\ref{tab:shot}). Its paper gives All F1 at one level
and each scale column at the best of its levels; the main comparison uses the single level, as published. A single
representation of its three levels combined scores 66.0 (Appendix~\ref{app:diag}).

\subsection{hBABEL}

Table~\ref{tab:hbabel} gives every column.

\begin{table}[!htbp]
\centering\small
\caption{hBABEL, the benchmark's evaluator. F1 of the frame-level behaviors averaged over the 10, 30, 60 and 90 most
frequent, and of the sequence-level behaviors over the 10, 30 and all 60; score: the mean of the two group means.
Published rows are from the hBehaveMAE paper~\citep{stoffl2024hbehavemae}, Table~4, with each score computed from its
columns.}\label{tab:hbabel}
\fitbox{\linewidth}{%
\begin{tabular}{@{}lcccccccc@{}}
\toprule
 & & \multicolumn{4}{c}{Frame-level, top} & \multicolumn{3}{c}{Sequence-level, top} \\
\cmidrule(lr){3-6}\cmidrule(lr){7-9}
Method & Score & 10 & 30 & 60 & 90 & 10 & 30 & 60 \\
\midrule
Generic recipe (ours), layout B & 20.9 & 34.1 & 20.9 & 14.1 & 11.8 & 32.1 & 17.7 & 14.7 \\
Hand-written variant (ours) & \textbf{22.1} & \textbf{34.9} & \textbf{22.3} & \textbf{14.8} & \textbf{12.4} & \textbf{33.7} & \textbf{19.8} & \textbf{15.9} \\
hBehaveMAE, released, level 2 (our run) & 16.5 & 29.8 & 17.9 & 11.6 & 10.0 & 22.9 & 13.9 & 10.5 \\
hBehaveMAE, best level for each column (our run) & 16.95 & 31.6 & 18.9 & 12.0 & 10.0 & 22.9 & 13.9 & 10.5 \\
BAMS, retrained (our run) & 14.5 & 29.4 & 16.7 & 10.5 & 8.0 & 19.4 & 11.5 & 7.5 \\
\midrule
hBehaveMAE, published & 16.98 & 31.6 & 18.9 & 12.0 & 9.7 & 22.8 & 14.0 & 10.9 \\
BehaveMAE ($15{\times}1{\times}15$ tokens), published & 17.3 & 34.0 & 20.3 & 13.2 & 10.8 & 22.3 & 13.4 & 9.5 \\
BehaveMAE ($15{\times}1{\times}75$ tokens), published & 16.5 & 30.2 & 17.7 & 11.4 & 9.2 & 23.4 & 14.0 & 10.4 \\
TVAE, published & 14.9 & 31.2 & 17.9 & 11.3 & 9.1 & 18.4 & 11.1 & 7.6 \\
TS2Vec, published & 14.4 & 29.6 & 17.3 & 10.9 & 8.8 & 17.5 & 10.8 & 7.9 \\
PCA, published & 13.8 & 27.7 & 16.2 & 10.4 & 8.4 & 17.5 & 10.5 & 7.5 \\
BAMS, published & 10.3 & 18.9 & 10.8 & 6.7 & 5.0 & 15.5 & 9.4 & 6.0 \\
\bottomrule
\end{tabular}}
\end{table}

\FloatBarrier
\section{Uncertainty and classifier checks}\label{app:robust}

\subsection{Primary comparisons and multiplicity}\label{app:multiplicity}

The primary comparisons are the 14 paired gaps in the bottom rows of Table~\ref{tab8:main}: the generic recipe minus
hBehaveMAE (four benchmarks), minus BAMS (three with a paired interval), minus HiDRA (two), and the hand-written variant
minus the generic recipe (five). The flies' gap to BAMS's published score has no paired interval and is left out. Every
other comparison in the paper is exploratory and reported without correction. Table~\ref{tab:holm} applies Holm's
step-down correction to the 14. On CalMS21 the $p$-values are the paired permutation tests of
Table~\ref{tab:generic_bca} (``$<$0.001'' taken as 0.001); on the other benchmarks $p$ is two-sided and read from the
10,000 paired resamples of each benchmark's interval script, rerun to keep them (they reproduce every published interval
exactly; the smallest possible $p$ is $10^{-4}$). Every gap whose interval excludes zero
survives at 0.05. With $p$ taken from the CalMS21 bootstrap itself instead of the permutation test, HiDRA's lead on
CalMS21 has an adjusted $p$ of 0.06 and every other reading is unchanged.

\paragraph{Encoder seeds and test videos together.} The intervals elsewhere hold each encoder's weights fixed. On
CalMS21, where hBehaveMAE has three training seeds (75.9, 75.2 and 73.8), resampling the seeds together with the 19
test videos (10,000 joint resamples) gives the generic recipe minus the mean over seeds as \gap{+4.8}{+2.7}{+8.5},
against [$+3.0$, $+8.4$] for the videos alone and [$+3.8$, $+5.9$] for the seeds alone. TS2Vec and CEBRA-Time on body-frame keypoints, retrained with two more seeds on CalMS21, score 50.0 to 53.4 and 69.7 to 70.6, and the same joint resampling gives the generic recipe minus each \gap{+28.1}{+23.6}{+36.7} and \gap{+9.6}{+5.3}{+19.3}. BAMS's four CalMS21 seeds are in Appendix~\ref{app:random}. The other encoders were trained
once, or are released checkpoints, so their intervals do not include training variance.

\paragraph{CalMS21 without attack.} Attack occurs in 5 of the 19 test videos, and a resample without an attack video
scores its F1 as zero. Scored as the macro F1 over investigation and mount only, the generic recipe leads hBehaveMAE by
\gap{+2.6}{+1.2}{+4.3} under the linear classifier and \gap{+3.6}{+2.0}{+5.7} under the fixed temporal head, hBehaveMAE on
body-frame keypoints by \gap{+2.8}{+1.4}{+4.5} and \gap{+1.2}{+0.3}{+2.7}, and BAMS by \gap{+21.6}{+17.5}{+28.7} under
the linear classifier; under the fixed temporal head BAMS is not resolved from it (\gap{-0.9}{-1.9}{+0.7}), and HiDRA leads it
under both (\gap{-5.2}{-6.8}{-4.0} and \gap{-4.4}{-5.9}{-3.3}).

\subsection{Random weights}\label{app:random}

Each random arm keeps the trained encoder's architecture, input and embedding script and replaces the trained weights
by the initialization of its authors' code. BAMS's arms (5 seeds) are embedded and read exactly as
the trained BAMS of each benchmark (Table~\ref{tab:random}); on CalMS21 trained BAMS has four seeds (0 to 3, the
authors' recipe with only the seed changed), which put the generic recipe \gap{+23.8}{+19.9}{+32.5} ahead under the
linear classifier and not resolved from it under the fixed temporal head (\gap{-0.1}{-1.5}{+1.5}), resampling seeds and videos. hBehaveMAE's arms (5 seeds on CalMS21, 3 on
Shot7M2) are embedded every 10th frame, which scores the trained models at 75.7 and 64.4 against the recorded 75.9 and
64.7, and their level is chosen on training labels. Paired gaps, weight seeds held fixed: on CalMS21 pretrained minus
random hBehaveMAE is \gap{+16.2}{+9.4}{+26.6} under the linear classifier and \gap{+13.9}{+8.7}{+21.8} under the fixed temporal
head; on Shot7M2 it is \gap{+4.3}{+4.1}{+4.6} and \gap{+2.7}{+2.5}{+2.8}, and random hBehaveMAE read by the fixed temporal head
is not resolved from the generic recipe (\gap{-0.1}{-0.5}{+0.2}).

\subsection{Bias-corrected intervals and permutation tests}

The CalMS21 Task 1 gaps are read with three kinds of uncertainty: the percentile interval used
throughout (10,000 resamples of the 19 test videos), the bias-corrected (BCa) interval from the same resamples, and the
two-sided $p$ of a paired permutation test that swaps the two methods' predictions video by video (10,000 swaps; the
smallest possible value is $10^{-4}$). The resampled gaps are right-skewed when a method's attack frames sit in a few
videos. For the generic recipe (Table~\ref{tab:generic_bca}):

\begin{table}[!htbp]
\centering\small
\caption{The generic recipe minus each CalMS21 representation, all labels: percentile interval, BCa interval and paired
permutation $p$ (each test video's two confusion matrices swapped at random, 10,000 times; two-sided), computed with the
script and draws used for the hand-written variant. $^\ast$Pretrained on pooled data that include
CalMS21.}\label{tab:generic_bca}
\setlength{\tabcolsep}{4pt}
\fitbox{\linewidth}{%
\begin{tabular}{@{}lccc@{}}
\toprule
Generic minus & Percentile & BCa & Permutation $p$ \\
\midrule
hBehaveMAE & \gap{+3.8}{+2.0}{+7.4} & [$+2.0$, $+7.4$] & $<$0.001 \\
BAMS & \gap{+23.2}{+19.4}{+31.8} & [$+12.7$, $+28.0$] & $<$0.001 \\
HiDRA, released$^\ast$ & \gap{-3.5}{-4.7}{-1.5} & [$-4.7$, $-1.3$] & $<$0.001 \\
LISBET, released & \gap{+4.9}{+2.8}{+9.3} & [$+2.1$, $+7.8$] & $<$0.001 \\
MARS's hand-built features & \gap{+1.0}{-1.1}{+8.8} & [$-1.9$, $+5.4$] & 0.5456 \\
hBehaveMAE, body-frame keypoints & \gap{+2.7}{+1.4}{+6.8} & [$+0.8$, $+4.8$] & 0.0026 \\
CEBRA-Time, body frame & \gap{+10.1}{+5.8}{+19.5} & [$+4.5$, $+17.0$] & $<$0.001 \\
TREBA, released & \gap{+17.7}{+12.9}{+28.6} & [$+11.7$, $+25.4$] & $<$0.001 \\
TS2Vec, body frame & \gap{+29.8}{+25.2}{+37.8} & [$+14.4$, $+34.8$] & $<$0.001 \\
Hand-written variant (ours) & \gap{-0.4}{-1.7}{+1.8} & [$-2.0$, $+1.3$] & 0.5930 \\
\bottomrule
\end{tabular}}
\end{table}

The bias-corrected intervals and permutation tests change no reading: the generic recipe stays ahead of hBehaveMAE, of BAMS and of LISBET, behind HiDRA, and not resolved from MARS's features or the hand-written variant.

\subsection{The temporal head's settings}\label{app:tunedhead}

The temporal head of Section~\ref{sec8:results} keeps its architecture (64 channels, kernel 5, dropout 0.1, weight decay
$10^{-4}$, 32 chunks of 256 frames) and chooses three settings for each representation separately: its context (dilations
1, 2, 4, 8 in four blocks, about $\pm$1\,s, or 1 to 32 in six blocks, about $\pm$4\,s), its training length (3,000 or 10,000
steps) and its learning rate ($10^{-3}$ or $3\cdot10^{-4}$). The choice uses training data only: on CalMS21 five-fold
cross-validation over the 70 training videos, and on Shot7M2 and hBABEL the evaluator's own split of its training clips,
with one head seed; the chosen setting is then trained with three seeds and scored once on the test set
(Table~\ref{tab:tunedhead}). Almost every representation chooses the longer context. The other temporal-head results in this
appendix, and the random-weight control, use one fixed setting for every method ($\pm$1\,s, 3,000 steps, $10^{-3}$), which we call the fixed temporal head.

\begin{table}[!htbp]
\centering\small
\caption{The temporal head's setting chosen on training data for each representation (context, steps, learning rate),
and its test score.}\label{tab:tunedhead}
\setlength{\tabcolsep}{3pt}
\fitbox{\linewidth}{%
\begin{tabular}{@{}llcc@{}}
\toprule
Benchmark & Representation & Setting & Test score \\
\midrule
CalMS21 & HiDRA & $\pm$4\,s, 3,000, $3\cdot10^{-4}$ & 88.1 \\
 & Raw keypoints & $\pm$4\,s, 3,000, $10^{-3}$ & 85.0 \\
 & MARS's features & $\pm$4\,s, 3,000, $3\cdot10^{-4}$ & 84.4 \\
 & LISBET & $\pm$4\,s, 3,000, $10^{-3}$ & 84.3 \\
 & BAMS & $\pm$4\,s, 3,000, $10^{-3}$ & 84.3 \\
 & Hand-written variant & $\pm$1\,s, 3,000, $3\cdot10^{-4}$ & 83.9 \\
 & Generic recipe & $\pm$4\,s, 3,000, $3\cdot10^{-4}$ & 83.4 \\
 & hBehaveMAE, body frame & $\pm$4\,s, 3,000, $3\cdot10^{-4}$ & 81.8 \\
 & CEBRA-Time, body frame & $\pm$4\,s, 3,000, $10^{-3}$ & 81.4 \\
 & TREBA & $\pm$4\,s, 10,000, $3\cdot10^{-4}$ & 80.7 \\
 & hBehaveMAE & $\pm$4\,s, 3,000, $3\cdot10^{-4}$ & 80.0 \\
 & TS2Vec, body frame & $\pm$4\,s, 3,000, $10^{-3}$ & 79.9 \\
Shot7M2 & Raw keypoints & $\pm$4\,s, 10,000, $10^{-3}$ & 83.1 \\
 & hBehaveMAE & $\pm$4\,s, 10,000, $10^{-3}$ & 82.6 \\
 & Hand-written variant & $\pm$4\,s, 10,000, $10^{-3}$ & 82.4 \\
 & Generic recipe & $\pm$4\,s, 10,000, $10^{-3}$ & 79.4 \\
 & BAMS & $\pm$4\,s, 10,000, $10^{-3}$ & 77.7 \\
hBABEL & Raw keypoints & $\pm$4\,s, 10,000, $10^{-3}$ & 23.0 \\
 & Generic recipe & $\pm$1\,s, 10,000, $10^{-3}$ & 22.7 \\
 & BAMS & $\pm$4\,s, 10,000, $10^{-3}$ & 21.4 \\
 & Hand-written variant & $\pm$1\,s, 3,000, $10^{-3}$ & 20.9 \\
 & hBehaveMAE & $\pm$1\,s, 10,000, $3\cdot10^{-4}$ & 16.7 \\
\bottomrule
\end{tabular}}
\end{table}

\subsection{Classifier checks}

\paragraph{A nonlinear probe.} Every result in the main tables uses the benchmarks' linear ridge classifier.
Table~\ref{tab:readouts} reads every CalMS21 Task 1 representation, in the variant the ridge chose, with one fixed
one-hidden-layer MLP instead (512 units on standardized inputs, a class-balanced loss, the same settings for every
method, none searched). Every method scores higher, and the readings hold: the generic recipe stays ahead of every
encoder but HiDRA, and not resolved from MARS's features or the hand-written variant (Table~\ref{tab:generic_cls}). The two
that gain most, TS2Vec and BAMS, still trail both variants by about 13 points. Read this way, our
BAMS (72.7) comes close to the 74.98 that MICE~\citep{su2026mice} reports for BAMS with a similar
probe, and our hBehaveMAE (82.4) is above MICE's 76.05. On the other benchmarks the same probe, in place of each evaluator's ridge classifier and with
everything else unchanged (the evaluator's training lists, its three training splits and its vote), raises every score
but the generic recipe's on the triplets (Table~\ref{tab:readouts_other}; the gaps to hBehaveMAE are in
Table~\ref{tab8:robust}). On Shot7M2 it lifts every scale of every method, and hBehaveMAE's actions and movemes most.

\paragraph{A temporal head.} A per-frame probe cannot integrate over time, which the window statistics do before any
classifier sees them. Table~\ref{tab:readouts} therefore also reads every CalMS21 Task~1 representation, in the variant
the ridge chose, with the fixed temporal head, one setting for every method (the head of Section~\ref{sec8:results} chooses its setting per representation; Appendix~\ref{app:tunedhead}): inputs standardized on the training frames, a
1$\times$1 convolution to 64 channels, four residual blocks of dilated convolutions (kernel 5, dilations 1, 2, 4 and 8,
a receptive field of 61 frames, about $\pm$1\,s), dropout 0.1 and a class-balanced loss, trained for 3,000 steps on
random 256-frame stretches of the 70 training videos with the same settings for every method, none searched, and scored
as the mean of three seeds. Every method gains, by 3.3 to 28.1 points, and the representations the ridge reads worst
gain most. On Shot7M2 and hBABEL the same head is trained inside each evaluator's
splits with one binary output per task and combined as the evaluator combines its ridge predictions (three seeds); on
hBABEL it scores slightly below the ridge for every method.

\begin{table}[!htbp]
\centering\small
\caption{CalMS21 Task 1 with all labels: macro F1 of every representation under the shared ridge classifier, one fixed
one-hidden-layer MLP probe and one fixed temporal convolutional head (mean of three seeds), and the hand-written variant
minus each with its paired 95\% interval under the two other readouts. The generic recipe's gaps are in
Table~\ref{tab:generic_cls}.}\label{tab:readouts}
\setlength{\tabcolsep}{4pt}
\fitbox{\linewidth}{%
\begin{tabular}{@{}lccccc@{}}
\toprule
 & \multicolumn{3}{c}{Macro F1} & \multicolumn{2}{c}{Hand-written minus it} \\
\cmidrule(lr){2-4}\cmidrule(lr){5-6}
Representation & Ridge & MLP probe & Fixed head & MLP probe & Fixed head \\
\midrule
Generic recipe (ours) & 79.7 & 85.8 & 83.0 & -- & -- \\
Hand-written variant (ours) & 80.1 & 85.8 & 84.5 & -- & -- \\
HiDRA, released & 83.2 & 88.0 & 88.4 & \gap{-2.2}{-3.4}{-0.5} & \gap{-4.0}{-5.2}{-2.7} \\
MARS's hand-built features & 78.7 & 85.3 & 84.0 & \gap{+0.5}{-0.5}{+3.4} & \gap{+0.4}{-1.0}{+3.6} \\
hBehaveMAE, body-frame keypoints & 77.0 & 83.9 & 81.0 & \gap{+1.9}{+0.8}{+3.4} & \gap{+3.5}{+2.0}{+6.7} \\
hBehaveMAE & 75.9 & 82.4 & 79.2 & \gap{+3.4}{+2.7}{+4.5} & \gap{+5.3}{+3.9}{+6.7} \\
LISBET, released & 74.9 & 79.9 & 83.3 & \gap{+5.9}{+4.6}{+8.5} & \gap{+1.1}{+0.1}{+2.5} \\
CEBRA-Time, body frame & 69.7 & 74.1 & 79.2 & \gap{+11.7}{+8.9}{+20.0} & \gap{+5.3}{+3.5}{+11.1} \\
TREBA, released & 62.0 & 68.2 & 78.1 & \gap{+17.6}{+15.1}{+24.9} & \gap{+6.4}{+4.9}{+8.4} \\
BAMS & 56.6 & 72.7 & 83.4 & \gap{+13.1}{+10.6}{+21.3} & \gap{+1.0}{+0.2}{+2.6} \\
TS2Vec, body frame & 50.0 & 72.6 & 78.1 & \gap{+13.2}{+11.0}{+21.1} & \gap{+6.4}{+4.3}{+12.8} \\
\bottomrule
\end{tabular}}
\end{table}

\begin{table}[!htbp]
\centering\small
\caption{CalMS21 Task 1 with all labels: the generic recipe minus each representation under the shared ridge, the fixed one-hidden-layer MLP probe and under the fixed temporal
convolutional head (mean of three seeds), with the settings above; and with the generic recipe's
windows and filters made past-only (the others unchanged). $^\ast$Pretrained on data that include CalMS21.}\label{tab:generic_cls}
\setlength{\tabcolsep}{3pt}
\fitbox{\linewidth}{%
\begin{tabular}{@{}lcccc@{}}
\toprule
Generic minus & Ridge & MLP probe & Fixed head & Past-only windows \\
\midrule

hBehaveMAE & \gap{+3.8}{+2.0}{+7.4} & \gap{+3.4}{+2.5}{+4.6} & \gap{+3.8}{+2.2}{+5.5} & \gap{-0.2}{-2.7}{+2.4} \\
BAMS & \gap{+23.2}{+19.4}{+31.8} & \gap{+13.1}{+10.6}{+21.4} & \gap{-0.4}{-1.2}{+0.9} & \gap{+19.2}{+16.0}{+25.3} \\
HiDRA, released$^\ast$ & \gap{-3.5}{-4.7}{-1.5} & \gap{-2.3}{-3.2}{-0.8} & \gap{-5.4}{-6.5}{-4.3} & \gap{-7.4}{-10.0}{-5.7} \\
LISBET, released & \gap{+4.9}{+2.8}{+9.3} & \gap{+5.9}{+4.5}{+8.5} & \gap{-0.3}{-1.5}{+1.1} & \gap{+0.9}{-1.1}{+3.7} \\
MARS's hand-built features & \gap{+1.0}{-1.1}{+8.8} & \gap{+0.4}{-0.9}{+3.4} & \gap{-1.0}{-2.0}{+1.5} & \gap{-3.0}{-5.4}{+2.5} \\
hBehaveMAE, body-frame keypoints & \gap{+2.7}{+1.4}{+6.8} & \gap{+1.9}{+1.0}{+3.2} & \gap{+2.0}{+0.9}{+4.2} & \gap{-1.3}{-2.9}{+0.8} \\
CEBRA-Time, body frame & \gap{+10.1}{+5.8}{+19.5} & \gap{+11.7}{+8.8}{+20.2} & \gap{+3.8}{+1.9}{+9.3} & \gap{+6.1}{+2.5}{+12.8} \\
TREBA, released & \gap{+17.7}{+12.9}{+28.6} & \gap{+17.6}{+15.0}{+24.8} & \gap{+4.9}{+3.0}{+6.9} & \gap{+13.7}{+8.8}{+22.1} \\
TS2Vec, body frame & \gap{+29.8}{+25.2}{+37.8} & \gap{+13.1}{+11.1}{+21.3} & \gap{+4.9}{+3.2}{+10.3} & \gap{+25.8}{+21.9}{+31.8} \\
Hand-written variant (ours) & \gap{-0.4}{-1.7}{+1.8} & \gap{-0.1}{-0.8}{+0.7} & \gap{-1.5}{-2.6}{-0.7} & \gap{-4.4}{-6.2}{-3.0} \\
\bottomrule
\end{tabular}}
\end{table}

\begin{table}[!htbp]
\centering\small
\caption{The other benchmarks under each readout, all labels, in each benchmark's own score: the evaluator's ridge
classifier, the fixed MLP probe and, on the frame-labeled human benchmarks, the fixed temporal head trained inside each
evaluator's splits (three seeds). hBehaveMAE at the level of the main comparison. Paired gaps to hBehaveMAE are in
Table~\ref{tab8:robust}.}\label{tab:readouts_other}
\fitbox{\linewidth}{%
\begin{tabular}{@{}llccc@{}}
\toprule
Benchmark & Representation & Ridge & MLP probe & Fixed head \\
\midrule
MABe22 triplets, clip-level F1 & Generic recipe (ours) & 83.1 & 82.8 & -- \\
 & Hand-written variant (ours) & 83.7 & 84.9 & -- \\
 & hBehaveMAE & 80.3 & 81.7 & -- \\
Shot7M2, All F1 & Generic recipe (ours) & 66.0 & 76.4 & 74.5 \\
 & Hand-written variant (ours) & 68.8 & 79.3 & 77.8 \\
 & hBehaveMAE & 64.7 & 80.1 & 77.1 \\
hBABEL, score & Generic recipe (ours) & 20.9 & 30.6 & 20.4 \\
 & Hand-written variant (ours) & 22.1 & 30.5 & 20.8 \\
 & hBehaveMAE & 16.5 & 24.7 & 16.1 \\
\bottomrule
\end{tabular}}
\end{table}

\paragraph{Standardized inputs.} The benchmark evaluators fit their ridge classifier on each representation as it is. A
ridge penalty treats every number alike, so a representation whose numbers differ widely in scale could be handicapped
by that protocol rather than by what it holds. Standardizing every number on the training frames first (CalMS21's 70
training videos; each evaluator's own training list elsewhere) changes no reading of either variant:
on CalMS21 no gap of the generic recipe moves by as much as a point, and the gaps to
hBehaveMAE on the other benchmarks do not change. We therefore keep the evaluators' protocol.

\paragraph{Past-only windows.} This check removes later frames from the recipe only. Since hBehaveMAE's embeddings, which it
is compared with, average over every 30\,s window that contains a frame and thus include later frames, this is a one-sided handicap rather than a
fairness check; it is run on CalMS21 with the linear classifier. Made past-only (trailing windows, a past-only median filter and backward-difference
velocities; the per-video scale of lengths still uses the whole video), the generic recipe scores 75.8 on CalMS21 with
all labels, 4.0 points below its centered windows, not resolved from hBehaveMAE and still ahead of BAMS,
CEBRA-Time, TREBA and TS2Vec (Table~\ref{tab:generic_cls}); its label curve is 27.7, 43.7, 66.2, 72.7, 75.0
and 75.8 at 2, 4, 8, 16, 35 and 70 videos, behind hBehaveMAE up to eight videos and not resolved from it from 16. The hand-written variant scores 78.7 on CalMS21 with all
labels, 1.4 points below the centered windows (\gap{-1.4}{-2.8}{-0.6}), still ahead of hBehaveMAE (\gap{+2.8}{+0.7}{+4.8}); with few labels (128 numbers per
frame) it trails hBehaveMAE with two and four labeled videos and leads from eight.

\FloatBarrier
\section{Few labels and smaller budgets}\label{app:fewlabels}

\begin{figure}[!htbp]
\centering
\includegraphics[width=0.72\linewidth]{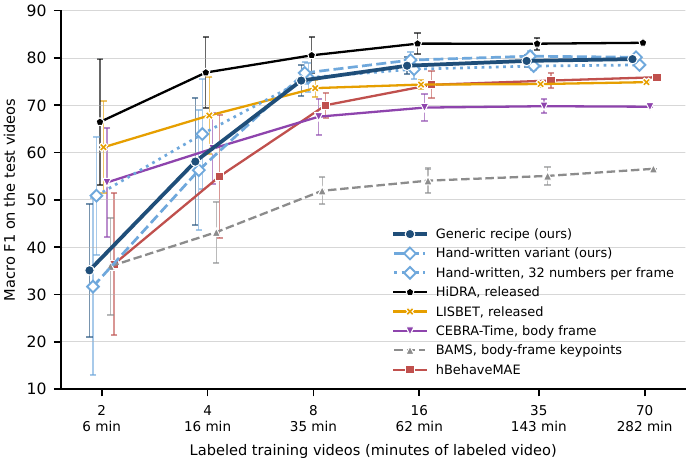}
\caption{CalMS21 Task 1 macro F1 against the number of labeled training videos (minutes of labeled video below each
count), for both variants, the hand-written variant cut to 32 numbers per frame, HiDRA, LISBET, CEBRA-Time on
body-frame keypoints, BAMS and hBehaveMAE; each point averages ten fixed sets of videos, and the error bars
show one standard deviation over the ten sets.}\label{fig:labels}
\end{figure}

In Table~\ref{tab:generic_curve} the generic recipe uses layout B on hBABEL, and in the rows with the budget chosen for each label amount, each encoder keeps the variant and
budget it chose in the same check of the hand-written variant.

\begin{table}[!htbp]
\centering\small
\caption{The generic recipe minus hBehaveMAE with fewer labels: CalMS21's ten fixed label sets per size, and five random
subsets of each other evaluator's training list (the same subsets for both methods); the column heads give CalMS21's
labeled videos, and the other benchmarks use the same share of their training lists (Figure~\ref{fig8:labels}).}\label{tab:generic_curve}
\setlength{\tabcolsep}{3pt}
\fitbox{\linewidth}{%
\begin{tabular}{@{}lcccccc@{}}
\toprule
Benchmark & 2 & 4 & 8 & 16 & 35 & 70 (all) \\
\midrule

CalMS21 & \gap{-1.3}{-6.9}{+3.2} & \gap{+3.1}{+0.7}{+5.1} & \gap{+5.2}{+3.9}{+7.2} & \gap{+4.0}{+2.6}{+6.3} & \gap{+4.2}{+2.5}{+7.5} & \gap{+3.8}{+2.0}{+7.4} \\
MABe22 triplets & \gap{-3.8}{-4.9}{-2.7} & \gap{-1.3}{-2.5}{-0.2} & \gap{+2.4}{+1.3}{+3.4} & \gap{+2.3}{+1.2}{+3.3} & \gap{+2.5}{+1.2}{+3.7} & \gap{+2.8}{+1.5}{+4.0} \\
Shot7M2 & \gap{-6.4}{-6.9}{-5.9} & \gap{-0.3}{-0.7}{+0.1} & \gap{+0.8}{+0.3}{+1.2} & \gap{+1.1}{+0.6}{+1.5} & \gap{+1.3}{+0.8}{+1.8} & \gap{+1.4}{+0.8}{+1.8} \\
hBABEL & \gap{+0.3}{+0.0}{+0.7} & \gap{+1.1}{+0.6}{+1.6} & \gap{+1.9}{+1.1}{+2.4} & \gap{+3.1}{+2.3}{+3.8} & \gap{+3.8}{+2.9}{+4.4} & \gap{+4.3}{+3.4}{+5.1} \\
\midrule
\multicolumn{7}{@{}l}{CalMS21, the generic recipe minus each further encoder} \\

LISBET & \gap{-26.0}{-30.0}{-21.2} & \gap{-9.7}{-11.7}{-7.8} & \gap{+1.6}{-0.1}{+3.9} & \gap{+4.0}{+1.8}{+7.4} & \gap{+4.9}{+2.8}{+9.2} & \gap{+4.9}{+2.8}{+9.3} \\
CEBRA-Time, body frame & \gap{-18.6}{-22.5}{-13.8} & \gap{-2.7}{-5.4}{+0.5} & \gap{+7.6}{+4.0}{+13.4} & \gap{+8.9}{+5.3}{+15.6} & \gap{+9.6}{+5.6}{+18.6} & \gap{+10.1}{+5.8}{+19.5} \\
BAMS & \gap{-0.9}{-4.0}{+2.3} & \gap{+14.9}{+12.7}{+17.4} & \gap{+23.3}{+20.4}{+27.9} & \gap{+24.3}{+21.0}{+30.2} & \gap{+24.3}{+21.0}{+31.7} & \gap{+23.2}{+19.4}{+31.8} \\
hBehaveMAE, body-frame keypoints & \gap{-6.3}{-10.8}{-2.1} & \gap{+5.0}{-0.6}{+10.3} & \gap{+5.3}{+2.2}{+9.8} & \gap{+3.3}{+1.5}{+6.6} & \gap{+3.5}{+1.8}{+7.9} & \gap{+2.7}{+1.4}{+6.8} \\
HiDRA, released & \gap{-31.4}{-36.4}{-26.6} & \gap{-18.8}{-23.1}{-15.3} & \gap{-5.3}{-7.3}{-3.9} & \gap{-4.7}{-6.4}{-3.4} & \gap{-3.6}{-5.1}{-1.8} & \gap{-3.5}{-4.7}{-1.5} \\
\midrule
\multicolumn{7}{@{}l}{CalMS21, budget chosen for each label amount by cross-validation (32, 64 or 128 numbers per frame)} \\

Budget chosen & 32 & 32 & 128 & 128 & 128 & 128 \\
Generic recipe & 41.8 & 60.7 & 75.2 & 78.4 & 79.4 & 79.7 \\
minus hBehaveMAE & \gap{-2.1}{-5.0}{+0.8} & \gap{+5.8}{+2.6}{+8.1} & \gap{+5.2}{+3.9}{+7.2} & \gap{+4.0}{+2.6}{+6.3} & \gap{+4.2}{+2.5}{+7.5} & \gap{+3.8}{+2.0}{+7.4} \\
minus LISBET & \gap{-19.3}{-21.7}{-16.6} & \gap{-7.1}{-10.2}{-4.0} & \gap{+1.6}{-0.1}{+3.9} & \gap{+4.0}{+1.8}{+7.4} & \gap{+4.9}{+2.8}{+9.2} & \gap{+4.9}{+2.8}{+9.3} \\
minus CEBRA-Time, body frame & \gap{-11.9}{-13.9}{-9.4} & \gap{-0.1}{-2.0}{+1.5} & \gap{+7.6}{+4.0}{+13.4} & \gap{+8.9}{+5.3}{+15.6} & \gap{+9.6}{+5.6}{+18.6} & \gap{+10.1}{+5.8}{+19.5} \\
minus HiDRA & \gap{-24.7}{-28.2}{-21.5} & \gap{-16.2}{-20.8}{-12.7} & \gap{-5.3}{-7.3}{-3.9} & \gap{-4.7}{-6.4}{-3.4} & \gap{-3.6}{-5.1}{-1.8} & \gap{-3.5}{-4.7}{-1.5} \\
\bottomrule
\end{tabular}}
\end{table}

\paragraph{LISBET and size.} LISBET, used through its authors' released CalMS21 embedder, scores 74.9 with all labels;
its own paper reports 70 for a classifier on its frozen embeddings~\citep{chindemi2023lisbet}. Below eight labeled videos
it leads the hand-written variant, as CEBRA-Time with body-frame input does; from eight
videos the variant leads it (\gap{+3.2}{+1.7}{+5.4} at eight). Against the generic recipe the same encoders lead only
below eight videos (Table~\ref{tab:generic_curve}). Much of LISBET's edge over the hand-written variant is size: LISBET gives 32 numbers per frame against
the variants' 128, and a linear classifier fitted on a couple of videos does better with fewer numbers. Cut to 32
numbers per frame (the first 16 components of each of its two PCAs), the hand-written variant scores 50.9 with two
labeled videos instead of 31.6, against LISBET's 61.1, and 78.5 with all labels, still ahead of LISBET
(\gap{+3.6}{+2.2}{+6.5}; permutation $p = 0.052$). Cutting every representation to 32 or
64 numbers in its own order of variance, the hand-written variant leads hBehaveMAE and CEBRA-Time on body-frame keypoints by 8 to 14 points with all labels, and is not resolved from HiDRA at 32 (\gap{-1.5}{-3.2}{+4.6}; bias-corrected $[-3.5, +2.9]$) but behind it at 64
(\gap{-2.4}{-3.6}{-0.2}). What LISBET encodes does not explain its edge: the readout gives it hBehaveMAE's
profile (Appendix~\ref{app:readouts}). The other behavior encoder, TREBA, is weak on its own: its released features are
made to be used next to the pose with a temporal classifier.

\paragraph{Budget chosen for each label amount.} With the budget (32, 64 or 128 numbers per frame) chosen for each
label amount by the same cross-validation on training videos, the generic recipe chooses 32 numbers with two and four
labeled videos and 128 from eight, is not resolved from hBehaveMAE with two videos and leads it from four
(Table~\ref{tab:generic_curve}).

\FloatBarrier
\section{What carries the result}\label{app:carries}

\subsection{Which windows}

Shot7M2's All F1 is split into
its three scales and hBABEL's score into its frame-level and sequence-level means. The encoder row is hBehaveMAE
(level~1 on Shot7M2, level~2 on hBABEL) and BAMS's published score on the flies.

\begin{table}[!htbp]
\centering\small
\caption{Which windows carry the result, by sub-scale, for both variants: each keeps only some of the windows,
with the benchmark's input, fit and evaluator, reduced with one PCA to at most the budget (the generic recipe from its
own layout's statistics; CalMS21's keep the histograms of its layout and have no whole-clip window; on the
flies, whose clips last 30\,s, the long set is the 8\,s window and the whole clip).}\label{tab:windows_both}
\setlength{\tabcolsep}{3pt}
\fitbox{\linewidth}{%
\begin{tabular}{@{}lcccccccccc@{}}
\toprule
 & CalMS21 & Triplets & Flies & \multicolumn{4}{c}{Shot7M2} & \multicolumn{3}{c}{hBABEL} \\
\cmidrule(lr){2-2}\cmidrule(lr){3-3}\cmidrule(lr){4-4}\cmidrule(lr){5-8}\cmidrule(lr){9-11}
Windows kept & macro F1 & clip & mean & All & activity & action & moveme & score & frame & sequence \\
\midrule
\multicolumn{11}{@{}l}{\emph{Generic recipe}} \\
Frame value only & 45.9 & 68.7 & 49.3 & 60.4 & 65.7 & 49.2 & 66.3 & 13.8 & 15.8 & 11.8 \\
Frame value, 0.5\,s and 2\,s & 82.0 & 71.1 & 52.6 & 62.7 & 72.4 & 55.9 & 59.9 & 17.6 & 20.4 & 14.9 \\
8\,s, 30\,s and whole clip & 66.7 & 82.9 & 49.8 & 59.7 & 92.8 & 47.1 & 39.1 & 19.9 & 17.3 & 22.6 \\
All (the recipe) & 79.7 & 83.1 & 52.4 & 66.0 & 90.4 & 55.5 & 52.2 & 20.9 & 20.2 & 21.5 \\
\multicolumn{11}{@{}l}{\emph{Hand-written variant}} \\
Frame value only & 68.3 & 70.2 & 45.9 & 60.0 & 67.4 & 49.7 & 62.8 & 15.2 & 18.1 & 12.4 \\
Frame value, 0.5\,s and 2\,s & 80.7 & 72.7 & 47.5 & 64.2 & 73.1 & 54.6 & 64.8 & 17.5 & 20.7 & 14.4 \\
8\,s, 30\,s and whole clip & 65.6 & 84.7 & 45.5 & 59.9 & 92.4 & 46.8 & 40.4 & 20.9 & 17.5 & 24.2 \\
All & 80.1 & 83.7 & 51.6 & 68.8 & 92.6 & 55.2 & 58.7 & 22.1 & 21.1 & 23.1 \\
\midrule
Encoder & 75.9 & 80.3 & 48.2 & 64.7 & 76.4 & 54.8 & 62.9 & 16.5 & 17.3 & 15.8 \\
\bottomrule
\end{tabular}}
\end{table}

Short windows carry Shot7M2's movemes and hBABEL's frame-level behaviors, and long windows its activities and hBABEL's
sequence-level behaviors (Table~\ref{tab:windows_both}). On CalMS21 the short windows alone score above the full set for
both variants (82.0 and 80.7 against 79.7 and 80.1), and on the triplets the long ones alone do for the hand-written
variant (84.7 against 83.7). The hand-written variant's short windows alone pass hBehaveMAE's best overall level on
Shot7M2's movemes, so its full set's lower moveme score reflects how it divides its 64 numbers between time scales. On
the triplets the lead comes from the long windows: without the whole-clip window the hand-written variant scores 77.9,
2.4 points below hBehaveMAE (\gap{-2.4}{-3.6}{-1.2}).

\subsection{The body frame and the relations between animals}

Table~\ref{tab:frames} gives the paired comparisons behind Section~\ref{sec8:needs}. The camera-coordinate representations keep
the windows, statistics, layout, budget and classifier of the body-frame ones and have no relations block; in
Figure~\ref{fig8:ladder}, CalMS21's camera-coordinate and body-frame points are in layout D (both mice's keypoints) and hBABEL's camera point in
layout C, while the generic recipe's are in its own layout B.
Shot7M2's camera coordinates are already centered on the pelvis horizontally, and hBABEL's have each frame's joint
centroid fixed; there the body frame scores 66.0 against 65.7 and 20.8 against 19.5, both in layout C. On CalMS21 the relations block is the generic recipe's only information about the other mouse; the
recipe without it again had its layout chosen by cross-validation (A 0.585, \textbf{B 0.657}, C 0.570, D 0.647). The
hand-written measurements split the same way (Table~\ref{tab8:rel}: relational 79.9, individual 64.2), and so do raw
keypoints in the hand-written variant's layout: the partner's keypoints and velocities in each mouse's own frame alone
score 79.2, and each mouse's own keypoints, velocities and turning alone 66.5 (Section~\ref{sec8:needs}'s 67.9, keypoints and
velocities in the generic recipe's own layout B, is the recipe without the block). Raw keypoints in the body frame do well on CalMS21 because, in that frame,
they say where the partner is.

\begin{table}[!htbp]
\centering\small
\caption{Paired comparisons for the body frame and for the relations between animals, all labels, in each benchmark's
own score. The raw-keypoint rows on CalMS21 use the hand-written variant's layout.}\label{tab:frames}
\fitbox{\linewidth}{%
\begin{tabular}{@{}llc@{}}
\toprule
Benchmark & Comparison & Gap \\
\midrule
CalMS21 & generic recipe minus the same without the relations block (67.9) & \gap{+11.9}{+9.1}{+15.2} \\
 & without the relations block minus hBehaveMAE & \gap{-8.0}{-11.5}{-3.5} \\
 & body frame minus camera coordinates, both mice's raw keypoints without the block & \gap{+20.7}{+16.3}{+27.0} \\
 & partner's raw keypoints in each mouse's frame minus hBehaveMAE & \gap{+3.2}{+1.0}{+4.8} \\
 & each mouse's own keypoints, velocities and turning minus hBehaveMAE & \gap{-9.5}{-13.3}{-4.6} \\
 & hand-written relational measurements minus hBehaveMAE & \gap{+4.0}{+1.8}{+5.5} \\
 & hand-written individual measurements minus hBehaveMAE & \gap{-11.8}{-14.5}{-9.6} \\
MABe22 triplets & generic recipe minus the same without the relations block & \gap{+0.6}{-0.1}{+1.3} \\
 & body frame minus camera coordinates, both without the block & \gap{+6.4}{+4.5}{+8.3} \\
 & without the relations block minus hBehaveMAE & \gap{+2.2}{+0.8}{+3.5} \\
 & hand-written individual measurements minus all & \gap{+0.3}{-0.5}{+1.1} \\
 & hand-written relational minus individual measurements & \gap{-15.3}{-17.3}{-13.2} \\
MABe22 flies & generic recipe minus the same without the relations block & \gap{-0.6}{-2.5}{+1.1} \\
 & generic recipe minus dish coordinates (which have no block) & \gap{+10.0}{+6.8}{+12.6} \\
 & hand-written: all minus individual measurements & \gap{+4.0}{+1.5}{+6.4} \\
 & hand-written relational minus individual measurements & \gap{-3.8}{-6.5}{-1.2} \\
\bottomrule
\end{tabular}}
\end{table}

\subsection{Generic time-series features}

MiniRocket and catch22 summarize the same body-frame keypoints (settings in Appendix~\ref{app:protocol}). On the three
benchmarks where we ran them they fall short of both variants, each gap resolved by a paired interval
(Table~\ref{tab:minirocket}), and both still beat hBehaveMAE on hBABEL; they were not run on the two mouse benchmarks.

\begin{table}[!htbp]
\centering\small
\caption{MiniRocket and catch22 on the same body-frame keypoints, all labels, in each benchmark's own score, and each
variant minus them with paired 95\% intervals.}\label{tab:minirocket}
\fitbox{\linewidth}{%
\begin{tabular}{@{}llccc@{}}
\toprule
Benchmark & Features & Score & Generic minus it & Hand-written minus it \\
\midrule
MABe22 flies & MiniRocket & 47.2 & \gap{+5.2}{+2.4}{+7.6} & \gap{+4.4}{+1.8}{+6.9} \\
 & catch22 & 42.7 & \gap{+9.7}{+6.6}{+12.3} & \gap{+8.9}{+6.2}{+11.3} \\
Shot7M2 & MiniRocket & 62.1 & \gap{+3.9}{+3.5}{+4.3} & \gap{+6.7}{+6.2}{+7.2} \\
 & catch22 & 60.7 & \gap{+5.4}{+4.9}{+5.8} & \gap{+8.2}{+7.7}{+8.6} \\
hBABEL & MiniRocket & 20.0 & \gap{+0.9}{+0.2}{+1.6} & \gap{+2.2}{+1.4}{+2.9} \\
 & catch22 & 19.8 & \gap{+1.1}{+0.3}{+1.6} & \gap{+2.3}{+1.6}{+2.9} \\
\bottomrule
\end{tabular}}
\end{table}

\FloatBarrier
\section{Why the encoders fall short: diagnostics and readouts}\label{app:diag}

\subsection{Windowed encoders and combined levels}

Given exactly our window statistics, reduction, budget and selection, hBehaveMAE stays well behind the hand-written
variant on CalMS21 and Shot7M2 (Table~\ref{tab:diag}). Part of that is the budget, since the windowed statistics hold
more than 128 numbers can keep: with 512 numbers per frame, or every statistic unreduced, windowed hBehaveMAE scores 77.2
and 77.0 on CalMS21, above its plain 75.9 but still behind both variants (Table~\ref{tab:diag_gaps}). At the budget the
windows lower hBehaveMAE on CalMS21, and on Shot7M2 they trade its movemes for activities.  On hBABEL and the
triplets we used hBehaveMAE's released embeddings, each benchmark's evaluator and its best windowed level on the test
set, a choice that favors it. Split this way, the hand-written variant's lead over hBehaveMAE comes mostly from the
windows on the mouse triplets (2.8 of 3.4 points), whose two labels describe whole clips, mostly from what each frame
carries on hBABEL (3.6 of 5.6), and entirely from it on CalMS21 and Shot7M2, where at the budget the windows cost
hBehaveMAE 2.2 and 2.3 points (with 512 numbers they lift it by 1.3 on CalMS21, so even there most of the lead, 2.9 of
4.2 points, comes from what each frame carries).

Concatenating hBehaveMAE's three released levels and compressing them to the budget, as its authors' own option does,
helps a little on Shot7M2 and hBABEL and hurts on the mouse triplets; the hand-written variant stays ahead on all three,
by 2.8 to 6.4 points, and the generic recipe equals the combined levels on Shot7M2 (66.0 each). Adding an
encoder's embedding to the measurements at the same budget does not improve on them, while adding the windowed
measurements to hBehaveMAE lifts it by 3.3 points on CalMS21 (to 79.2, still below the hand-written variant alone).

\begin{table}[!htbp]
\centering\small
\caption{Diagnostics for the hand-written variant, all labels: each encoder plain, through our window statistics, and
with its embedding added to our windowed measurements at the same budget (top), and hBehaveMAE's three levels combined
(bottom). The generic recipe's
counterparts are in Table~\ref{tab8:diag}.}\label{tab:diag}
\setlength{\tabcolsep}{3pt}
\fitbox{\linewidth}{%
\begin{tabular}{@{}lcccc@{}}
\toprule
Encoder & Plain & + our windows & + measurements & Hand-written \\
\midrule
\multicolumn{5}{@{}l}{\emph{CalMS21 Task 1}} \\
hBehaveMAE & 75.9 & 73.8 & 79.2 & \textbf{80.1} \\
\multicolumn{5}{@{}l}{\emph{Shot7M2}} \\
hBehaveMAE (best level) & 64.7 & 62.3 & 68.0 & \textbf{68.8} \\
\multicolumn{5}{@{}l}{\emph{MABe22 mouse triplets}} \\
hBehaveMAE (best level) & 80.3 & 83.1 & -- & \textbf{83.7} \\
\multicolumn{5}{@{}l}{\emph{hBABEL}} \\
hBehaveMAE (best level) & 16.5 & 18.5 & -- & \textbf{22.1} \\
\midrule
 & Best level & Levels combined & & Hand-written \\
\midrule
MABe22 mouse triplets & 80.3 & 77.3 & & \textbf{83.7} \\
Shot7M2 & 64.7 & 66.0 & & \textbf{68.8} \\
hBABEL & 16.5 & 17.3 & & \textbf{22.1} \\
\bottomrule
\end{tabular}}
\end{table}

\begin{table}[!htbp]
\centering\small
\caption{Paired gaps of the diagnostics, all labels, in each benchmark's own score. Added: the encoder's embedding added
to our windowed measurements at the same budget.}\label{tab:diag_gaps}
\fitbox{\linewidth}{%
\begin{tabular}{@{}llc@{}}
\toprule
Benchmark & Comparison & Gap \\
\midrule
CalMS21 & hand-written minus windowed hBehaveMAE & \gap{+6.4}{+4.6}{+8.1} \\
 & hand-written minus windowed hBehaveMAE, 512 numbers & \gap{+2.9}{+1.5}{+4.0} \\
 & hand-written minus windowed hBehaveMAE, every statistic (1,152 numbers) & \gap{+3.1}{+2.3}{+4.1} \\
 & hand-written at 512 numbers minus windowed hBehaveMAE at 512 & \gap{+3.6}{+2.2}{+4.7} \\
 & generic recipe minus windowed hBehaveMAE, 512 numbers (77.2) & \gap{+2.5}{+1.4}{+4.4} \\
 & generic recipe minus windowed hBehaveMAE, unreduced (77.0) & \gap{+2.7}{+1.3}{+5.0} \\
 & windowed minus plain hBehaveMAE & \gap{-2.2}{-3.3}{-0.8} \\
 & hand-written with hBehaveMAE added, minus hand-written & \gap{-0.9}{-2.0}{+0.5} \\
 & hand-written with hBehaveMAE added, minus hBehaveMAE & \gap{+3.3}{+1.2}{+6.4} \\
MABe22 triplets & hand-written minus windowed hBehaveMAE & \gap{+0.5}{-1.0}{+2.0} \\
 & windowed minus plain hBehaveMAE & \gap{+2.8}{+2.0}{+3.7} \\
 & hBehaveMAE's levels combined minus its best level & \gap{-3.0}{-3.8}{-2.2} \\
 & hand-written minus levels combined & \gap{+6.4}{+5.0}{+7.8} \\
 & hand-written without the whole-clip window minus hBehaveMAE & \gap{-2.4}{-3.6}{-1.2} \\
 & hand-written without the whole-clip window minus all & \gap{-5.8}{-7.0}{-4.6} \\
Shot7M2 & hand-written minus windowed hBehaveMAE & \gap{+6.5}{+6.1}{+6.9} \\
 & windowed minus plain hBehaveMAE & \gap{-2.3}{-2.8}{-1.9} \\
 & levels combined minus best level & \gap{+1.3}{+1.0}{+1.6} \\
 & hand-written minus levels combined & \gap{+2.8}{+2.4}{+3.3} \\
 & hand-written with hBehaveMAE added, minus hand-written & \gap{-0.9}{-1.2}{-0.5} \\
hBABEL & hand-written minus windowed hBehaveMAE & \gap{+3.6}{+2.8}{+4.4} \\
 & windowed minus plain hBehaveMAE & \gap{+1.9}{+1.2}{+2.6} \\
 & levels combined minus best level & \gap{+0.7}{+0.4}{+1.0} \\
 & hand-written minus levels combined & \gap{+4.9}{+4.0}{+5.6} \\
\bottomrule
\end{tabular}}
\end{table}

\subsection{Supplying the measurements}\label{app:supply}

Appending the hand-written measurements' frame values to hBehaveMAE's embedding, without the windows and keeping every
number of both, helps it on all four benchmarks tested, and the motion measurements least (Table~\ref{tab:supply_gaps}; on CalMS21 this
comparison splits the 23 measurements into 14 of motion and 9 others, the budget-matched one below into 10 and 13): 0.1
to 1.4 points for motion and 0.6 to 3.1 for the others, ahead of motion on all four (Table~\ref{tab:supply_gaps}). On
CalMS21 the others, mostly relational (the distance between the mice, contact, each nose to the other mouse, facing),
close most of the gap to the hand-written variant (79.0 against 80.1); elsewhere they add little. Motion helps least
frame by frame because it is noisy at the scale of a frame; what the labels use is motion summarized over seconds.
What lifts hBehaveMAE is the measurements with their windows, whereas windowing its own embedding lifts it far less.

Appending also adds numbers beyond the budget. At the budget (Table~\ref{tab:supply_bm}), hBehaveMAE is first reduced
by a whitened PCA, fitted without labels on training frames only (CalMS21's 70 training videos; the evaluator's training
clips elsewhere), to the budget minus the number of appended measurements (10 motion or 13 other measurements of 128 numbers on CalMS21, 27 or 58 on the triplets, and 16 or 24 of 64 on Shot7M2 and hBABEL), which keeps 96 to 99.9\% of its variance.
Most of the appended gain then disappears: motion adds at most 0.3 points and costs 0.8 on the triplets, and the other
measurements add 2.9 on CalMS21, where they are mostly relational, and between $-1.6$ and $+0.4$ elsewhere. Frame values at the budget therefore do not close
hBehaveMAE's gap to the recipe, in line with the reading that the labels use measurements summarized over windows.

\begin{table}[!htbp]
\centering\small
\caption{The measurements' frame values appended to hBehaveMAE's embedding, all labels: paired gaps in each benchmark's own score.}\label{tab:supply_gaps}
\fitbox{\linewidth}{%
\begin{tabular}{@{}llc@{}}
\toprule
Benchmark & Comparison & Gap \\
\midrule
CalMS21 & hBehaveMAE with the motion frame values minus hBehaveMAE & \gap{+0.8}{+0.4}{+1.5} \\
 & hBehaveMAE with the other frame values minus hBehaveMAE & \gap{+3.1}{+2.4}{+3.8} \\
 & with motion minus with the other frame values & \gap{-2.3}{-3.1}{-1.4} \\
MABe22 triplets & hBehaveMAE with the motion frame values minus hBehaveMAE & \gap{+0.1}{+0.0}{+0.2} \\
 & hBehaveMAE with the other frame values minus hBehaveMAE & \gap{+0.6}{+0.2}{+1.1} \\
 & with motion minus with the other frame values & \gap{-0.5}{-1.0}{-0.1} \\
Shot7M2 & hBehaveMAE with the motion frame values minus hBehaveMAE & \gap{+1.4}{+1.2}{+1.4} \\
 & hBehaveMAE with the other frame values minus hBehaveMAE & \gap{+1.8}{+1.6}{+1.9} \\
 & with motion minus with the other frame values & \gap{-0.4}{-0.6}{-0.3} \\
hBABEL & hBehaveMAE with the motion frame values minus hBehaveMAE & \gap{+0.7}{+0.5}{+0.8} \\
 & hBehaveMAE with the other frame values minus hBehaveMAE & \gap{+1.5}{+1.2}{+1.7} \\
 & with motion minus with the other frame values & \gap{-0.8}{-1.0}{-0.5} \\
\bottomrule
\end{tabular}}
\end{table}

\subsection{What the representations encode}\label{app:readouts}

For the four benchmarks with hBehaveMAE, a ridge regression reads each hand-written measurement out of each
representation; its $R^2$ on held-out clips (five-fold cross-validation, no behavior label) measures how much of that
quantity the representation makes linearly available, which is what the benchmarks' linear classifiers can use. The
hand-written variant is built from these measurements and recovers them largely, though not fully after PCA to the
budget. Motion averages speeds, approach, activity and angular rates; angles and shape average heading, facing, joint
angles and body shape; positions average distances, positions and heights. For hBehaveMAE each bar of
Figure~\ref{fig8:encode} takes the best of its three levels, and negative values are drawn as zero.  Table~\ref{tab:generic_encode} gives the values
behind Figure~\ref{fig8:encode}.

\begin{table}[!htbp]
\centering\small
\caption{Held-out $R^2$ of a ridge readout of the hand-written measurements from each representation (the
regression of Section~\ref{sec8:why}, on the same frames; Figure~\ref{fig8:encode}): frame values averaged within each kind, and all 30\,s means.
hBehaveMAE: its best level for each column.}\label{tab:generic_encode}
\setlength{\tabcolsep}{4pt}
\fitbox{\linewidth}{%
\begin{tabular}{@{}llcccc@{}}
\toprule
Benchmark & Representation & Positions & Motion & Angles, shape & 30\,s means \\
\midrule

CalMS21 & hBehaveMAE (best level) & 0.86 & 0.26 & 0.33 & 0.43 \\
 & Generic recipe (ours) & 0.94 & 0.45 & 0.50 & 0.75 \\
 & Hand-written variant (ours) & 0.97 & 0.72 & 0.69 & 0.94 \\
\addlinespace
MABe22 triplets & hBehaveMAE (best level) & 0.75 & 0.31 & 0.24 & 0.51 \\
 & Generic recipe (ours) & 0.66 & 0.46 & 0.36 & 0.58 \\
 & Hand-written variant (ours) & 0.94 & 0.70 & 0.79 & 0.86 \\
\addlinespace
Shot7M2 & hBehaveMAE (best level) & 0.96 & 0.83 & 0.92 & 0.70 \\
 & Generic recipe (ours) & 0.79 & 0.68 & 0.66 & 0.79 \\
 & Hand-written variant (ours) & 0.89 & 0.85 & 0.84 & 0.84 \\
\addlinespace
hBABEL & hBehaveMAE (best level) & 0.61 & 0.19 & 0.52 & 0.32 \\
 & Generic recipe (ours) & 0.58 & 0.42 & 0.55 & 0.65 \\
 & Hand-written variant (ours) & 0.85 & 0.70 & 0.83 & 0.79 \\
\bottomrule
\end{tabular}}
\end{table}

Table~\ref{tab:further_encode} applies Section~\ref{sec8:why}'s linear readout, on the same CalMS21 training frames
and with no behavior label, to every further encoder: held-out $R^2$ averaged over the measurements of each kind (their
frame values), and over all measurements' 30\,s means. The first hBehaveMAE row is its level 2, the level used on CalMS21
and in the rank correlations; the second takes its best level for each column, as Figure~\ref{fig8:encode} does. Section~\ref{sec8:why}'s rank correlations use
these values before rounding; rounded, some of them tie. Across the ten representations the rank correlations of the
scores are 0.82 with motion ($p = 0.006$), 0.78 with the 30\,s means ($p = 0.01$), 0.65 with angles and shape
($p = 0.05$), 0.38 with positions and 0.56 with the relational measurements' frame values: motion and the angles and
shape track the ranking, positions do not clearly.

\paragraph{Dependent points.} Three encoders (hBehaveMAE, TS2Vec and CEBRA-Time) appear with two inputs, so the ten
points are not independent, and the encoders also differ in architecture, input frame and size. Recomputed from the
two-decimal values of Table~\ref{tab:further_encode} (on which the ten-point correlation is 0.83, exact permutation
$p = 0.004$), the correlation with motion is 0.86 with one point per encoder (each two-input encoder as the mean of its
two versions; exact $p = 0.024$ over the seven), 0.89 with the camera-input versions ($p = 0.012$) and 0.70 with the
body-frame versions ($p = 0.09$), and the partial rank correlation given the input frame is 0.80 (permutation
$p = 0.011$). All $p$ here and in Section~\ref{sec8:why} are two-sided. Within a single input frame only three or four
encoders remain, too few to test. The association survives these checks, but it remains a correlation across
representations that differ in several ways at once.

\begin{table}[!htbp]
\centering\small
\caption{Linear readout of the hand-written measurements from every CalMS21 representation (held-out
$R^2$): frame values averaged within each kind, and all 30\,s means.}\label{tab:further_encode}
\fitbox{\linewidth}{%
\begin{tabular}{@{}lcccc@{}}
\toprule
Representation & Positions & Motion & Angles, shape & 30\,s means \\
\midrule
hBehaveMAE, level 2 & 0.79 & 0.23 & 0.28 & 0.43 \\
hBehaveMAE, best level per column & 0.86 & 0.26 & 0.33 & 0.43 \\
TS2Vec, camera & 0.71 & 0.06 & 0.10 & 0.17 \\
CEBRA-Time, camera & 0.85 & 0.09 & 0.18 & 0.17 \\
TS2Vec, body frame & 0.68 & 0.11 & 0.29 & 0.22 \\
CEBRA-Time, body frame & 0.93 & 0.14 & 0.30 & 0.22 \\
BAMS & 0.77 & 0.15 & 0.37 & 0.17 \\
hBehaveMAE, body-frame keypoints & 0.81 & 0.20 & 0.34 & 0.44 \\
TREBA, released & 0.57 & 0.26 & 0.22 & 0.12 \\
LISBET, released & 0.78 & 0.20 & 0.28 & 0.28 \\
HiDRA, released & 0.81 & 0.40 & 0.59 & 0.32 \\
Generic recipe (ours) & 0.94 & 0.45 & 0.50 & 0.75 \\
Hand-written variant (ours) & 0.97 & 0.72 & 0.69 & 0.94 \\
\bottomrule
\end{tabular}}
\end{table}

Two further tests on CalMS21 ask whether the motion is in the embeddings in a
form a linear readout cannot use, and whether the encoders' other content gets in the way (Table~\ref{tab:why}; every
frame-10 sample of the training videos, five-fold cross-validation over videos). Some encoders given camera coordinates
spend most of their variance on where the body parts are in the image: the raw keypoints, linearly, explain 78\% of
hBehaveMAE's level 0 (Table~\ref{tab:why} shows level 2, the one used on CalMS21) and 66\% of TS2Vec's, while the
measurements explain at most 6\% of any encoder given camera coordinates. A one-hidden-layer MLP reads little more motion out of the encoders than the ridge does (at most 0.10 more $R^2$ for
any encoder in Table~\ref{tab:why}; hBehaveMAE's level 2 from 0.23 to 0.25, LISBET from 0.20 to 0.28), and from every encoder trained on keypoints at most 0.44, against 0.72
to 0.84 for the hand-written variant, so their motion is not recoverable by a small MLP
either, though a larger readout might recover more. Removing each mouse's own position and heading from every embedding, by its linear prediction from
those, and re-scoring never helps (at most $+0.0$ points) and mostly hurts: the position content is not in the
classifier's way, and it uses some of it. The hand-written variant changes on many time scales at once (median 4.3\,s, middle half
1.1 to 7.5\,s), most encoders on a narrow band (medians of 1.5 to 3.6\,s in Table~\ref{tab:why}, middle halves at most 1.2\,s wide;
BAMS, TREBA and HiDRA change much faster), and behavior bouts are short (median 1.1\,s for
investigation, 1.5\,s for attack, 2.0\,s for mount). The encoders do not lose because they are crowded with where the animals are.

\begin{table}[!htbp]
\centering\small
\caption{What each CalMS21 representation holds, how much motion a linear and a nonlinear readout get out of it, how
fast it changes, and what removing each mouse's position and heading does to its macro F1 with all labels.}\label{tab:why}
\setlength{\tabcolsep}{3.5pt}
\fitbox{\linewidth}{%
\begin{tabular}{@{}lcccccc@{}}
\toprule
 & \multicolumn{2}{c}{Explained by ($R^2$)} & \multicolumn{2}{c}{Motion read out ($R^2$)} & & \\
\cmidrule(lr){2-3}\cmidrule(lr){4-5}
Representation & image keypoints & measurements & linear & MLP & Time scale, s & Position removed \\
\midrule
Hand-written variant (ours) & $-0.01$ & 0.34 & 0.72 & 0.84 & 4.3 & $-0.8$ \\
hBehaveMAE (level 2) & 0.07 & 0.06 & 0.23 & 0.25 & 1.9 & $-1.9$ \\
LISBET, released & 0.00 & 0.40 & 0.20 & 0.28 & 2.2 & $-1.3$ \\
TREBA, released & 0.00 & 0.18 & 0.26 & 0.34 & 0.5 & $-0.1$ \\
TS2Vec, camera & 0.66 & 0.02 & 0.06 & 0.12 & 3.0 & $-3.9$ \\
CEBRA-Time, camera & 0.20 & 0.06 & 0.09 & 0.13 & 3.6 & $+0.0$ \\
TS2Vec, body frame & 0.00 & 0.30 & 0.11 & 0.21 & 2.1 & $-2.7$ \\
CEBRA-Time, body frame & $-0.01$ & 0.28 & 0.14 & 0.18 & 2.3 & $-0.4$ \\
BAMS & 0.00 & 0.13 & 0.15 & 0.24 & 0.1 & $-2.4$ \\
hBehaveMAE, body-frame keypoints & $-0.01$ & 0.07 & 0.20 & 0.19 & 1.5 & -- \\
HiDRA, released & 0.08 & 0.09 & 0.40 & 0.44 & 0.4 & $-0.6$ \\
\bottomrule
\end{tabular}}
\end{table}

\FloatBarrier
\section{Further encoders on CalMS21}\label{app:further}

CalMS21's unlabeled videos and released weights let many encoders be run on it. Are BAMS and hBehaveMAE simply the
wrong encoders? We ran five further
self-supervised encoders through the same protocol (Table~\ref{tab:further}): two general-purpose ones,
TS2Vec~\citep{yue2022ts2vec}, the standard self-supervised encoder for time series, and CEBRA-Time~\citep{schneider2023cebra},
the contrastive embedding most used for behavioral and neural data; and three built for mouse behavior,
TREBA~\citep{sun2021treba}, from CalMS21's own authors, LISBET~\citep{chindemi2023lisbet}, a transformer trained on social
tasks, and HiDRA~\citep{marks2026doom}, the self-supervised trunk of an annotator built from a 2025 community challenge.
We trained the general-purpose encoders with their authors' code and settings on each benchmark's own training data (on
CalMS21 with its unlabeled videos), given the keypoint coordinates each benchmark provides or, on CalMS21, also both
mice's keypoints in the resident's frame. TREBA, LISBET and HiDRA are their authors' released features, embedder and
trunks.

\begin{table}[!htbp]
\centering\small
\caption{Further encoders, and MARS's hand-built features, with all labels, in each benchmark's own score, with the
shared ridge classifier.}\label{tab:further}
\setlength{\tabcolsep}{5pt}
\fitbox{\linewidth}{%
\begin{tabular}{@{}lccccc@{}}
\toprule
Encoder & CalMS21 & Mouse triplets & Fly groups & Shot7M2 & hBABEL \\
\midrule
\multicolumn{6}{@{}l}{\emph{General-purpose encoders (time series, neural data)}} \\
TS2Vec & 43.8 & 59.3 & 39.5 & 56.6 & 15.2 \\
CEBRA-Time & 49.5 & 64.9 & 38.2 & 57.5 & 11.4 \\
TS2Vec, body frame & 50.0 & -- & -- & -- & -- \\
CEBRA-Time, body frame & 69.7 & -- & -- & -- & -- \\
\multicolumn{6}{@{}l}{\emph{Encoders built for mouse behavior}} \\
TREBA, released & 62.0 & -- & -- & -- & -- \\
LISBET, released & 74.9 & -- & -- & -- & -- \\
HiDRA, released & \textbf{83.2} & 75.8 & -- & -- & -- \\
\multicolumn{6}{@{}l}{\emph{Encoders built for behavior in general (main comparison)}} \\
BAMS & 56.6 & -- & -- & 57.3 & 14.5 \\
hBehaveMAE & 75.9 & 80.3 & -- & 64.7 & 16.5 \\
hBehaveMAE, body-frame keypoints & 77.0 & -- & -- & -- & -- \\
\midrule
\multicolumn{6}{@{}l}{\emph{Training-free}} \\
Generic recipe (ours) & 79.7 & 83.1 & \textbf{52.4} & 66.0 & 20.9 \\
Hand-written variant (ours) & 80.1 & \textbf{83.7} & 51.6 & \textbf{68.8} & \textbf{22.1} \\
Hand-written, 32 numbers per frame & 78.5 & -- & -- & -- & -- \\
MARS's hand-built features & 78.7 & -- & -- & -- & -- \\
\bottomrule
\end{tabular}}
\end{table}

\paragraph{Another hand-built set.} MARS~\citep{segalin2021mars} computes its own hand-designed pose
features for exactly these recordings. Run through the same protocol (the 148 features per mouse of its released
code, without its 10 pixel features, which need the videos; its own windowing over 0.1 to 0.7\,s; our reduction to
128 numbers and the shared classifier), they score 78.7 with all labels, not resolved from either variant (the hand-written variant minus them
\gap{+1.4}{-0.7}{+7.7}, the generic recipe \gap{+1.0}{-1.1}{+8.8}), ahead of CEBRA-Time and not resolved from hBehaveMAE or LISBET
(MARS's features minus hBehaveMAE \gap{+2.8}{-3.4}{+5.5}, minus LISBET \gap{+3.8}{-0.7}{+6.7}, minus CEBRA-Time \gap{+9.0}{+5.8}{+13.3}). MARS's set thus scores between and around our variants, though none of these gaps is resolved, and the 87.8 that \citet{li2026visualcontext} report for MARS's features with a trained recurrent
classifier fits the same picture.

\paragraph{General-purpose encoders do worse still.} TS2Vec and CEBRA-Time fall below both behavior encoders on
CalMS21 (43.8 and 49.5), and both variants lead them by more than 30 points with all labels. Training TS2Vec ten times
longer did not help on the training videos. On the other four benchmarks, given the coordinates each benchmark provides,
both fall far behind both variants and behind hBehaveMAE wherever it was run (Table~\ref{tab:further}), even though
each keeps the better of two training variants on the test set (TS2Vec at 600 or 6,000 iterations, CEBRA-Time at offset
10 or 36), which favors it.

\paragraph{Body-frame input for the encoders.} Measuring each animal in its own frame is much of both
variants' advantage (Section~\ref{sec8:needs}). Given the same body-frame input, CEBRA-Time and TS2Vec gain, but none catches
either variant (Table~\ref{tab:further}). CEBRA-Time gains 20 points (to 69.7) and TS2Vec 6 (to 50.0). hBehaveMAE gains a little: 77.0 against 75.9 with all labels, an
unresolved gain (\gap{+1.1}{-1.9}{+3.4}) smaller than its seed-to-seed spread (75.9, 75.2 and 73.8), and 5 points with
two labeled videos (\gap{+5.0}{+2.0}{+7.5}). The generic recipe stays ahead of each (Table~\ref{tab:generic_cls}), and
the hand-written variant leads hBehaveMAE on body-frame keypoints by 3.1 (\gap{+3.1}{+1.6}{+5.9}) and the others by 10
to 30 points. The frame thus explains part of the encoders' deficit, not all of it; what they still miss is mostly
motion (Appendix~\ref{app:readouts}).

\paragraph{One encoder beats both variants on CalMS21.} HiDRA~\citep{marks2026doom} is the newest self-supervised mouse
encoder with released code and weights: a
keypoint-trajectory forecaster with message passing within and between the two mice and a recurrent layer at each of
its four levels. We use its released ensemble of five trunks (140 million parameters), pretrained on 9,008 videos from
the 15 laboratories of its benchmark and three public datasets, CalMS21, CRIM13 and MABe22 (its model card), possibly
including their test videos, which likely favors it; its embedding is reduced to the benchmark's budget. On CalMS21 it beats both variants: 83.2 against 79.7 and 80.1
with all labels (\gap{-3.5}{-4.7}{-1.5} and \gap{-3.1}{-4.8}{-1.6}, each variant minus HiDRA), and it leads at every label budget
(Figure~\ref{fig:labels}), on investigation and mount more than on attack (Table~\ref{tab:perbehavior}). Its lead does not carry over to the other mouse benchmark. On the MABe22 mouse triplets (each clip run for the three pairs of
mice, since HiDRA reads mice in pairs, and each mouse averaged over its two pairs) it scores 75.8 on the two clip-level
tasks, behind both variants (\gap{+7.3}{+5.9}{+8.6} and \gap{+7.9}{+6.6}{+9.2}, each variant minus HiDRA) and behind the
released hBehaveMAE (80.3); on the frame-level behaviors it scores 21.3 (Table~\ref{tab:mabe22}).

HiDRA fits the account of Section~\ref{sec8:why}: trained to forecast where the keypoints go rather than to
reconstruct where they are, it carries more motion and more of the body's angles and shape than any other encoder
trained on keypoints, though less than the hand-written variant, and fewer of the 30\,s means than hBehaveMAE
(Table~\ref{tab:further_encode}). Which of its parts gives it the edge is left to future work.
\FloatBarrier

\end{document}